\documentclass{article}

\usepackage{iclr2021_conference,times}

\usepackage{amsmath,amsfonts,bm}

\def\eqref#1{equation~\ref{#1}}
\def\1{\bm{1}}

\DeclareMathAlphabet{\mathsfit}{\encodingdefault}{\sfdefault}{m}{sl}
\SetMathAlphabet{\mathsfit}{bold}{\encodingdefault}{\sfdefault}{bx}{n}

\DeclareMathOperator*{\argmax}{arg\,max}

\usepackage{hyperref}
\usepackage{url}
\usepackage{booktabs}       % professional-quality tables
\usepackage{amsfonts}       % blackboard math symbols
\usepackage{nicefrac}       % compact symbols for 1/2, etc.
\usepackage{microtype}      % microtypography
\usepackage{xcolor}         % colors

\usepackage[ruled,vlined]{algorithm2e}
\usepackage{mathtools}
\usepackage{color, colortbl}
\definecolor{Gray}{gray}{0.9}
\usepackage{tikz}
\usetikzlibrary{arrows.meta,calc,decorations.markings,math,arrows.meta}
\usepackage{subcaption}
\usepackage{multirow}

\makeatletter
\newcommand*{\centerfloat}{%
  \parindent \z@
  \leftskip \z@ \@plus 1fil \@minus \textwidth
  \rightskip\leftskip
  \parfillskip \z@skip}
\makeatother

\title{Quantifying Local Specialization in Deep Neural Networks}

\author{\textbf{Shlomi Hod,$^{*1,3}$ Daniel Filan,$^{*1,2}$} Stephen Casper,$^{*1,4}$\\\textbf{Andrew Critch, $^{1,2}$ Stuart Russell $^{1,2}$}\\
$^1$Center for Human-Compatible AI (CHAI)\\$^2$University of California Berkeley\\$^3$Boston University\\$^4$MIT Computer Science and Artificial Intelligence Laboratory (CSAIL)\\$*$ Equal contribution\\
\texttt{shlomi@bu.edu}\hspace{8pt} \texttt{daniel\_filan@berkeley.edu}\hspace{8pt}\texttt{scasper@csail.mit.edu}}

\iclrfinalcopy

\begin{document}

\maketitle

\begin{abstract}
A neural network is locally specialized to the extent that parts of its computational graph (i.e.\ structure) can be abstractly represented as performing some comprehensible sub-task relevant to the overall task (i.e.\ functionality).
Are modern deep neural networks locally specialized?
How can this be quantified?
In this paper, we consider the problem of taking a neural network whose neurons are partitioned into clusters, and quantifying how functionally specialized the clusters are.
We propose two proxies for this: \emph{importance}, which reflects how crucial sets of neurons are to network performance; and \emph{coherence}, which reflects how consistently their neurons associate with features of the inputs.
To measure these proxies, we develop a set of statistical methods based on techniques conventionally used to interpret individual neurons.
We apply the proxies to partitionings generated by spectrally clustering a graph representation of the network's neurons with edges determined either by network weights or correlations of activations.
We show that these partitionings, even ones based only on weights (i.e.\ strictly from non-runtime analysis), reveal groups of neurons that are important and coherent.
These results suggest that graph-based partitioning can reveal local specialization and that statistical methods can be used to automatedly screen for sets of neurons that can be understood abstractly. Code is available at \href{https://github.com/thestephencasper/local_specialization}{https://github.com/thestephencasper/local\_specialization}.
\end{abstract}

\section{Introduction} \label{sec:introduction}

Modularity is a common property of complex systems, both natural and artificial~\citep{clune2013evolutionary, baldwin2000design, booch2007object}. 
The ability for a system to separate different sub-tasks into distinct architectural components has benefits such as intelligibility and adaptivity.
Therefore, it would be valuable to determine the extent to which neural networks are \emph{locally specialized}: that is, how much their functionality can be abstracted into comprehensible sub-tasks, each localized to different groups of neurons.
Existing work~\citep{filan2021clusterability} has produced methods of partitioning a network into connected groups of neurons.
In this paper, we focus on developing quantifiable proxies for local specialization.
We then apply them, together with variants of these partitioning methods, to a variety of image classification networks.

There exists a body of research for developing networks which either have distinct architectural building blocks \citep{alet2018modular, parascandolo2018learning, goyal2019recurrent} or are trained in a way that promotes modularity via regularization or parameter isolation \citep{kirsch2018modular, de2019continual, filan2021clusterability}.
Yet in machine learning, it is more common to encounter networks whose architecture and training are not designed to separate the computation of sub-tasks. 
For example, in computer vision, networks are generally trained end-to-end with all of the filters in one layer connected to all filters in the next.
Do such networks nonetheless exhibit local specialization?
There is some evidence for this.
For example, by methodical manual investigation, \citet{cammarata2020thread} discovered
sub-networks which perform human-explainable sub-tasks such as car detection via neurons which detect different car parts.
More scalably detecting local specialization in networks would help us to extend our understanding of their learning dynamics and to expand our interpretability toolbox by suggesting an additional level of abstraction beyond single-neuron methods.

In this paper, we systematically analyze the extent to which networks which are not explicitly trained to be modular nonetheless exhibit local specialization.
First, this requires a method for breaking down a network's computational graph into groups of neurons.
For this, we use spectral clustering on a graph representation of the network, using an extension of the methods of~\citet{filan2021clusterability}.
Second, this requires a scalable method for approximating the degree to which a partitioning shows local specialization.
We do this by applying interpretability tools to clusters of neurons as a way of quantifying proxies for local specialization.

\textbf{Proxies for local specialization:}
Our definition of local specialization requires human-comprehensible sub-tasks relevant to the overall task to be localized in particular subsets of neurons.
However, directly determining this would require a human in the loop, making it difficult to scale such a method.
So: how could measuring local specialization be automated?
Consider an idealized prototype of a highly modular network that has subsets of neurons performing sub-tasks in which
    (1) each sub-task is necessary for high performance on the overall task,
    (2) each sub-task is implemented by a single subset only,
    and (3) each subset executes only a single sub-task.
The combination of (1) and (2) suggests that the removal of one of the subsets from the network would harm performance because the network lacks the implementation of a necessary sub-task that is localized to the subset.
We say that such a subset is \emph{important}.
Next, given that neurons are frequently understood as feature detectors,
(3) suggests that the neurons in a subset should tend to be strongly activated by inputs which contain features relevant to the subset's sub-task.
We say that such a subset is \emph{coherent}.
Figure~\ref{fig:halves_vis} of Appendix~\ref{app:correlation_based_visualization} provides an illustrative example of coherence in networks trained on a task that lends itself to parallel processing of sub-tasks (though this is not one of our key experiments presented in Section~\ref{sec:experiments}).
Measuring importance and coherence thus offers a sense of the degree to which a partitioning of a network's neurons contains subsets that meet these prototypical conditions, despite not perfectly satisfying them.

\textbf{Results and contributions:}
To measure these proxies,
we take partitionings of neurons in a network generated by spectral clustering,
and analyze them using methods from the interpretability literature that have conventionally been applied to single neurons.
Our key results are shown in Tables \ref{tab:fisher_stats} and \ref{tab:effect_sizes}.
We find that these partitions have groups of neurons that are disproportionately likely to be important compared to random groups of neurons, but that are not always more important than random groups of neurons on average.
We also find that the groups of neurons in the partitionings are reliably more coherent than random ones, though only with respect to features other than class label.

By showing that our partitioning methods are able to reveal local specialization, these results suggest that they can be
used to screen for interesting, abstractable subsets of units and better understand deep networks.
Our key contributions are threefold:
\begin{enumerate}
    \item Introducing two proxies, importance and coherence, to assess whether a partitioning of a network shows local specialization and identify which subsets of neurons are the most responsible.
    \item Quantifying these proxies by applying single-neuron interpretability methods on subsets of neurons in an automated fashion.
    \item Applying our methods on the partitions produced by spectral clustering on a range of neural networks and finding evidence of local specialization captured by these partitions.
\end{enumerate}

Code is available at \href{https://github.com/thestephencasper/local_specialization}{https://github.com/thestephencasper/local\_specialization}.

\section{Related Work} \label{sec:related_works}

The work most closely related to ours is \citet{filan2021clusterability} who also use spectral clustering to establish that deep networks are often clusterable and investigates what factors influence clusterability. 
They also introduce two methods for regularization for clusterability among clusters of neurons. 
We extend their work by bridging graphical clusterability
and local specialization.
This line of work inherits insights from network science involving clustering in general \citep{girvan2002community, newman2004finding}, and spectral clustering \citep{shi2000normalized, von2007tutorial} in particular.

In our experiments we combine clustering with interpretability tools to measure importance and coherence. 
We use neural lesions \citep{zhou2018revisiting} and feature visualization \citep{olah2017feature, watanabe2019interpreting}, but in a similar way, other interpretability techniques including analysis of selectivity \citep{morcos2018importance, madan2020capability}, network ``dissection'' \citep{bau2017network, mu2020compositional}, earth-mover distance \citep{testolin2020deep}, or intersection information \citep{panzeri2017cracking} could also be combined with clustering-based partitionings under a similar framework. 
Relatedly, \citet{cammarata2020thread} demonstrate that feature visualization and analysis of weights can be used to identify groups of neurons whose functionality is human-interpretable.

This work adds to a body of research focused on modularity and compositionality in neural systems either at the neuron level
\citep{you2020graph, mu2020compositional, voss2021branch} or at the subnetwork level \citep{lake2015human, lake2017building, csordasneural2020, udrescu2020ai}. 
There also exist techniques for developing more modular networks which either have an explicitly modular architecture \citep{alet2018modular, parascandolo2018learning, goyal2019recurrent} or are trained in a way that promotes modularity via regularization or parameter isolation \citep{kirsch2018modular, de2019continual}.

\section{Methods} \label{sec:methods}

\begin{figure}[t!]
    \centering
    \includegraphics[width=0.95\columnwidth]{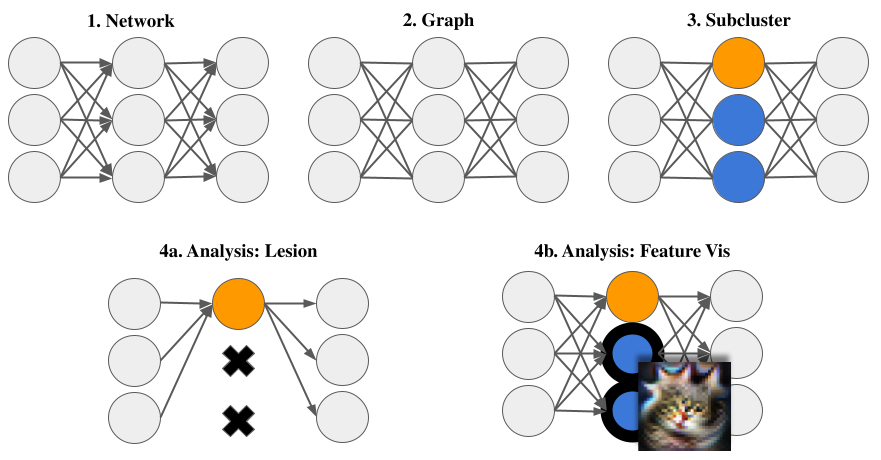}
    \caption{\textbf{Our procedural pipeline.} The first three steps generate a partitioning of the network into ``subclusters'' which we analyze using (4a) lesion and (4b) feature visualization methods to measure importance and coherence compared to random subclusters. Finally, (not shown in the pipeline), we aggregate results to produce Fisher statistics, $p$ values, and effect measures. These final steps are shown in Figure~\ref{fig:schematic-extension}.}
    \label{fig:schematic}
\end{figure}

To evaluate local specialization, our procedural pipeline is as follows.
(1) We begin with a trained neural network;
(2) construct a graph from it, treating each neuron as a node;
(3) perform spectral clustering on the graph to obtain a partitioning or ``clustering'' of neurons which we then further divide by layer to obtain a ``subclustering'';
(4) use our proxies for local specialization to analyze the subclusters, operationalized by lesioning neurons or feature visualization and comparing them to random subclusters;
and (5) aggregate results across the network to obtain a $p$ value, effect measure, and a quantity which we refer to as the \emph{Fisher statistic}. This pipeline is outlined in Figure~\ref{fig:schematic} and Figure~\ref{fig:schematic-extension}, and each step is explained in detail below. 

\subsection{Generating Partitions with Spectral Clustering} \label{sec:partition}

To partition a network into clusters, we use an approach based on \citet{filan2021clusterability} which consisted of two steps: \emph{``graphification''} - transforming the network into an undirected, edge-weighted graph; and \emph{clustering} - obtaining a partitioning via spectral clustering.

\textbf{Graphification:} To perform spectral clustering, a network must be represented as an undirected graph with non-negative edges. 
For MLPs (multilayer perceptrons), each graph node corresponds to a neuron in the network including input and output neurons. 
For CNNs (convolutional neural networks), a node corresponds to a single channel (which we also refer to as a ``neuron'') in a convolutional layer.\footnote{
If a unit is used as inputs to multiple layers, as happens in ResNets with skip connections, we consider these inputs to be separate neurons.
}
For CNNs, we ignore input, output, and fully-connected layers when clustering. 

For graphification, we test two ways of assigning adjacency edges between neurons: with weights and with correlations. 
For weight-based clustering with dense layers, if two neurons have a weight connecting them in the network, their corresponding vertices are connected by an edge with weight equal to the absolute value of the network's weight between the neurons. 
For convolutional channels, we connect them by an edge with weight equal to the $L_1$ norm for the corresponding 2D kernel slice. 
If layers are connected but with a batch-normalization layer in between, we mimic the scaling performed by the batch norm operation by multiplying weights by $\gamma / \sqrt{\sigma^2 + \varepsilon}$ where $\gamma$ is the scaling factor, $\sigma$ is the moving standard deviation, and $\varepsilon$ is a small constant. 
Notably, this method of constructing the graph requires no dataset or runtime analysis of the network.

While graphification via weights only results in connections between neurons in adjacent layers, doing so with correlations creates more dense graphs. 
With this method, we connect the nodes for two neurons with their squared Spearman correlation across a validation set. 
Spearman correlation gives the Pearson (linear) correlation between ranks and reflects how well two sets of data can be related by a monotonically increasing function.\footnote{
We use Spearman rather than Pearson correlation because networks are nonlinear, and there is no particular reason to expect associations between arbitrary neurons to be linear.} 
Rather than using neurons' post-ReLU outputs to calculate these correlations, we use their pre-ReLU activations with the goal of extracting richer data from them.\footnote{Although all negative values are mapped to zero by a ReLU, we expect the degree of negativity to carry information about the possible presence (or lack thereof) of features which the neuron was meant to detect, especially for networks trained with dropout.}
Again, for convolutional channels, we take the $L_1$ norm of activations before calculating Spearman correlations.
The fact that we take absolute valued weights or squared correlations to construct edges between neurons in graphification means that our analysis does not discriminate between positive and negative associations between neurons. 

In addition to graphification via weights versus activations, we also test two scopes with which to perform clustering:
network-wide and layer-wise.
For network-wide clustering, we cluster on one graph for the network as a whole. 
For layer-wise clustering, we produce a partitioning for each layer $l$ individually by clustering on the graph of connections between $l$ and the layers adjacent to $l$.
Ultimately, we run 4 sets of experiments on each network by clustering \{weights, activations\} $\times$ \{network-wide, layer-wise\}.

\textbf{Spectral (sub)clustering:} We perform normalized spectral clustering \citep{shi2000normalized} on the resulting graphs to obtain a partition of the neurons into clusters. 
For all sub-ImageNet experiments, we set the number of clusters to $k=16$, while for ImageNet-scale networks, we use $k=32$.
In Appendix \ref{app:alternate_k}, we reproduce a subset of sub-ImageNet experiments with $k \in \{8, 12\}$ showing that results are robust to alternate choices of $k$.
Refer to Appendix~\ref{app:spectral_clustering_algorithm} for a complete description of the spectral clustering algorithm. 

Layers at different depths of a network tend to develop different representations.
Therefore, for network-wide clustering in which clusters of neurons span more than one layer, we analyze clusters one layer at a time. 
We call these sets of neurons within the same cluster and layer \emph{subclusters}.
To ensure comparability between these clusterings when performing layer-wise clustering, we set the number of clusters per layer to be the same as the number produced in that layer with network-wide clustering. 
In our experiments, we compare these subclusters to other random sets of neurons of the same size in the same layer. 
We refer to subclusters identified by the clustering algorithm as ``true subclusters'' and sets of random neurons as ``random subclusters.''
Random subclusters form the natural control condition to test whether the specific partitioning of neurons exhibits importance or coherence compared to alternative partitions, while taking account of layer and size.

\subsection{Analysis Pipeline} \label{sec:statistical_analysis}

\textbf{Overview:} In our experiments, we measure the degree to which true subclusters identified via spectral clustering are more important and coherent than random subclusters of the same size and layer.
Operationalizations are given in subsections~\ref{sec:lesion} and \ref{sec:feature_visualization}, but in brief, the measure for importance of a subcluster quantifies the performance reduction from dropping out
the neurons in that subcluster, and the measures for coherence quantify the degree to which the neurons in a subcluster are mutually associated with some feature of an input.
For each subcluster with more than one neuron and which does not include every neuron in the layer, we calculate a measure of importance or coherence and compare it to that of 19 random subclusters.
These experiments and measures are discussed in detail in Section~\ref{sec:experiments}.
We present two measures of how true and random subclusters compare under these proxies. 
First, we calculate a measure of whether true subclusters are disproportionately often more important or coherent than random subclusters, called the Fisher statistic.
This value is our primary focus. 
However, for additional resolution, we also calculate an effect measure used to assess the importance and coherence of the `typical' subcluster.
Importantly, the Fisher statistic and effect measure quantify different things.

\begin{figure}[t]
    \centering
    \caption{\textbf{Illustration of Fisher statistics of various percentile distributions.}
    A VGG network trained on CIFAR-10 is partitioned using four methods (\{weights, activations\} $\times$ \{network-wide, layer-wise\})
    and analyzed for coherence (see discussion of visualization scores in Section~\ref{sec:feature_visualization}) to produce the collection of percentiles for each subcluster.
    This figure shows histograms of the percentile distribution for each clustering, and their associated Fisher statistics.
    Recall that a lower percentile means that a true subcluster is more coherent than random subclusters while controlling for layer and size.
    The activation-based clusterings have disproportionately many low percentiles, and Fisher statistics greater than 2.
    Table~\ref{tab:fisher_stats} shows that these trends are statistically significant when aggregated over five models.}
    \label{fig:hist-precentiles}
    \includegraphics[width=\columnwidth]{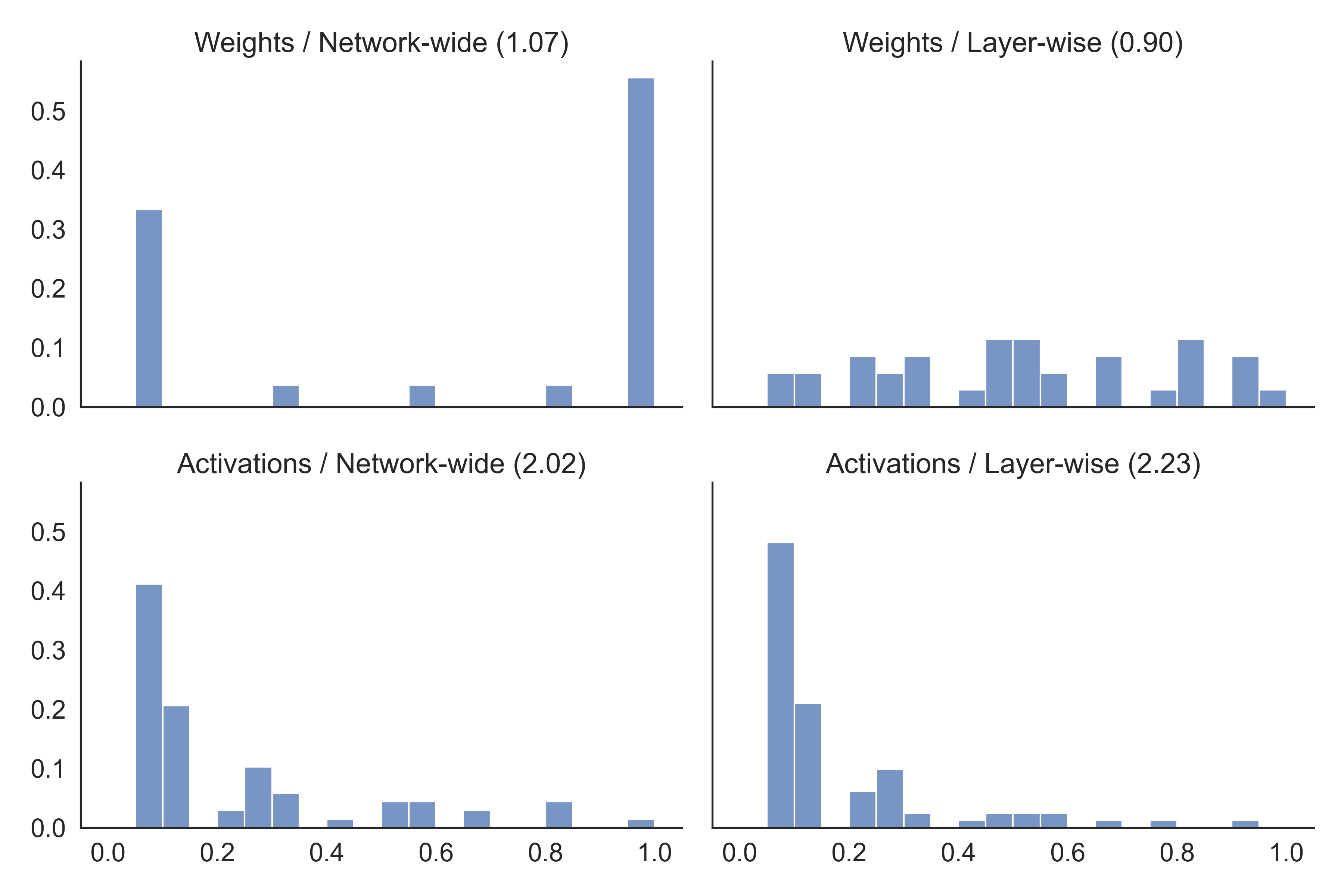}
    
\end{figure}

\textbf{Fisher statistics:}
We wish to test whether spectral clustering methods find subsets of neurons which satisfy our proxies for local specialization more than if we had simply chosen random subsets.
To do this, for each subcluster measurement we take the percentile of each true subcluster relative to the distribution of measurements of random subclusters.\footnote{For metrics where high values indicate local specialization, we take the percentile of the negative metric, so that low percentiles consistently indicate local specialization.}
Next, we use the Fisher method to test whether the subclusters in a single network satisfy our proxies more than random subsets of neurons.
To do so, we first center the subcluster percentiles around 0.5, which under the null hypothesis would give a granular, unbiased approximation of the uniform distribution.
We then combine the centered percentiles $\{p_1, \dotsc, p_n\}$ into the Fisher statistic $(-1/n) \sum_{i=1}^n \log p_i$.
For reference, the Fisher statistic of a uniform distribution of percentiles in our setting is 0.98.
Figure~\ref{fig:hist-precentiles} shows example distributions of percentiles and their Fisher statistics.
Note also that since the $\log$ function has a larger derivative near 0 than near 1, low percentiles have greater influence on the Fisher statistic, so J- or U-shaped distributions can also have Fisher statistics greater than 1, even if there are more high percentiles than low percentiles.
For all non-ImageNet architectures, we train and analyze 5 different networks per condition and report the mean Fisher statistic.

The Fisher statistic of $n$ uniformly-distributed random percentiles multiplied by $2n$ takes a chi-squared distribution with $2n$ degrees of freedom.
This lets us produce a $p$ value for each network, testing whether this statistic is higher than the null hypothesis would produce, which would mean that there were more low percentiles than if subclusters were distributed uniformly.\footnote{The fact that we coarsely measure percentiles and then center them makes this test conservative because our statistic is more sensitive to low percentiles than high percentiles.}
This procedure is illustrated in Figure~\ref{fig:schematic-extension}.
For non-ImageNet architectures, we need to aggregate the five $p$ values we get per condition, one for each network.
To do so, in each condition, we take the mean of the $p$ values
and correct it using the corresponding quantile of a $\textrm{Bates}(n=5)$ distribution\footnote{The $\textrm{Bates}(n)$ quantile function is $F_n(x) = \frac{1}{n!}\sum_{k=0}^{\lfloor nx \rfloor} (-1)^k \binom{n}{k}(nx-k)^{n-1}$ \citep{marengo2017geometric}.} which gives the distribution of the mean of 5 independent random variables uniformly distributed on $[0,1]$.\footnote{We use the Bates method for aggregation here instead of using the Fisher method because in this case, all five $p$ values are from identically configured experiments, and the Bates test is less sensitive to low outliers.}

This produces a $p$ value for every network architecture, partitioning method, and local specialization proxy.
We next correct for multiple testing using the Benjamini Hochberg method \citep{benjamini1995controlling}, controlling the false discovery rate at the $\alpha=0.05$ level.
See Appendix~\ref{app:benjamini_hochberg} for details.
Table~\ref{tab:fisher_stats} shows in bold the Fisher statistics (or means thereof) that are statistically significantly greater than 0.98.

In summary, (1) we aggregate each network's subcluster percentiles using the Fisher method, yielding a Fisher statistic and a $p$ value, (2) we aggregate the Fisher statistics and $p$ values across identically-configured replicates of the same experiment by taking the mean and using the Bates method, respectively, and (3) we correct for multiple comparisons using the Benjamini-Hochberg procedure.

\textbf{Effect measures:} In addition to $p$ values, we also calculate effect measures which give a sense of how different results are for true and random subclusters are \emph{on average}. 
The effect measure is the mean over subclusters of $2x/(x+\mu)$, where $x$ is a true subcluster measure of importance/coherence and $\mu$ is the mean over that of random subclusters.\footnote{If $x$ is ever less than 0 for a subcluster, we conservatively replace it with 0.}
We do this as opposed to simply taking $x/\mu$ to avoid division by zero.
This results in effect measures in the interval $[0,2]$, and which side of 1 they are on indicates whether the true subclusters are more important/coherent than random ones.
For ease of interpretation, note that if $2x/(x+\mu) = 1 + y$, then $x/\mu \approx 1 + 2y$ for $y \ll 1$, so an effect measure of $1.05$ would mean that the measure of a true subcluster was $\approx 10\%$ higher than the expected measure of a random subcluster.
Together with these effect measures, we also report their standard errors.

\textbf{Differences between Fisher statistics and effect measures:}
In some of our experiments, the Fisher statistics and effect measures seem to ``disagree'' with one suggesting that the network was locally specialized and the other suggesting it was not.
This potential for disagreement is due to the fact that Fisher statistics are based on the percentiles of subcluster measurements relative to the distribution of those of random subclusters,
while effect measures compare the value of subcluster measurements to the mean value of random subcluster measurements.
When the Fisher statistic seems to indicate local specialization but the effect measure does not, this means that on the relevant metric, there are more subclusters than would be expected under the null hypothesis whose metric value is higher than that of random subclusters, but the ``typical'' subcluster has a metric value similar to or less than an average random subcluster.
In other words, the partitioning method has detected some subclusters that satisfy our proxies for local specialization, but the typical subcluster found does not.
This is compatible with a J- or U-shaped distribution of subcluster percentiles. 
We consider this a positive result, indicating that our partitioning method is detecting some local specialization.

\section{Experiments} \label{sec:experiments}

To show the applicability of our methods at different scales, we experiment with a range of networks. 
For small-scale experiments, we train MLPs with 4 hidden layers of 256 neurons each and small convolutional networks with 3 layers each of 64 neurons followed by a dense layer of 128 neurons trained on the MNIST \citep{lecun1998gradient} dataset. 
At a mid scale, we train VGG-style CNNs containing 13 convolutional layers using the architectures of~\citet{simonyan2014very} trained on CIFAR-10 \citep{krizhevsky2009learning} using the procedure of \citet{DBLP:conf/acpr/LiuD15}, which includes weight decay and dropout for regularization.
Finally, for ImageNet \citep{krizhevsky2009learning} scale, we analyze pretrained ResNet18 \citep{he2016deep}, VGG-16, and VGG-19 \citep{simonyan2014very} models. 
Further details including hyperparameters and test performances are in Appendix~\ref{app:network_training_details}.

\begin{table}[h!]
\footnotesize
\caption{Fisher statistics (or means over 5 runs) for (1) lesion-based experiments measuring importance via overall accuracy drops (\textcolor{blue}{Acc. Drop}) and coherence via the class-wise range of accuracy drops (\textcolor{orange}{Class Range}); and (2) Feature visualization-based experiments in networks measuring coherence via the optimization score (\textcolor{orange}{Vis Score}) and the entropy of network outputs (\textcolor{orange}{Softmax $H$}). Each row corresponds to a network paired with a partitioning method. Fisher statistics above 0.98 indicate that subclusters
satisfy our local specialization proxies disproportionately more than random subclusters do.
Values statistically significantly greater than 0.98 are \textbf{bolded}. Section~\ref{sec:statistical_analysis} details the calculation of these statistics and their $p$ values.
}
\label{tab:fisher_stats}
\centerfloat
\begin{tabular}{l l r r r r }
\toprule
 & &  \multicolumn{2}{c}{\textbf{Lesion}}
& \multicolumn{2}{c}{\textbf{Feature Visualization}}\\

\textbf{Network} & \textbf{Partitioning}
& {\textbf{\textcolor{blue}{Acc.\ Drop}}}
& {\textbf{\textcolor{orange}{Class Range}}}
& {\textbf{\textcolor{orange}{Vis Score}}}
& {\textbf{\textcolor{orange}{Softmax $\mathbf{H}$}}} \\
\midrule

\multirow{4}{*}{MLP, MNIST} & Weight/Network & $\mathbf{2.13}$ & $0.91$ & $\mathbf{1.32}$ & $0.92$ \\
& Weight/Layer & $\mathbf{2.05}$ & $0.91$ & $\mathbf{1.21}$ & $\mathbf{1.23}$ \\
& Act./Network & $\mathbf{1.46}$ & $1.00$ & $\mathbf{1.34}$ & $\mathbf{1.15}$ \\
& Act./Layer & $\mathbf{1.69}$ & $1.03$ & $\mathbf{1.36}$ & $\mathbf{1.12}$ \\ \hline

\multirow{4}{*}{CNN, MNIST} & Weight/Network & $\mathbf{1.29}$ & $0.84$ & $1.10$ & $0.93$ \\
& Weight/Layer & $1.10$ & $0.94$ & $1.02$ & $0.99$ \\
& Act./Network & $\mathbf{1.73}$ & $0.70$ & $1.09$ & $0.90$  \\
& Act./Layer & $\mathbf{1.46}$ & $0.92$ & $1.05$ & $0.98$ \\ \hline

\multirow{4}{*}{VGG, CIFAR-10} & Weight/Network & $\mathbf{1.50}$ & $\mathbf{2.12}$ & $\mathbf{1.46}$ & $0.99$ \\
& Weight/Layer & $\mathbf{1.15}$ & $\mathbf{1.27}$ & $1.00$ & $0.97$ \\
& Act./Network & $\mathbf{1.40}$ & $0.97$ & $\mathbf{2.34}$ & $1.08$ \\
& Act./Layer & $\mathbf{1.56}$ & $1.03$ & $\mathbf{2.67}$ & $\mathbf{1.12}$ \\ \hline

\multirow{4}{*}{VGG-16, ImageNet} & Weight/Network & $\mathbf{2.54}$ & $0.49$ & $\mathbf{1.72}$ & $1.19$ \\
& Weight/Layer & $\mathbf{2.15}$ & $0.56$ & $\mathbf{1.90}$ & $1.06$ \\
& Act./Network & $\mathbf{1.89}$ & $0.63$ & $\mathbf{1.82}$ & $1.07$  \\
& Act./Layer & $\mathbf{1.66}$ & $0.70$ & $\mathbf{1.85}$ & $0.98$ \\ \hline

\multirow{4}{*}{VGG-19, ImageNet} & Weight/Network & & & $\mathbf{1.91}$ & $1.03$ \\
& Weight/Layer & & & $\mathbf{2.23}$ & $1.00$ \\
& Act./Network & & & $\mathbf{1.87}$ & $1.10$ \\
& Act./Layer & & & $\mathbf{2.01}$ & $0.98$ \\ \hline

\multirow{4}{*}{ResNet18, ImageNet} & Weight/Network & $\mathbf{1.42}$ & $1.13$ & & \\
& Weight/Layer & $\mathbf{1.29}$ & $0.99$ & & \\
& Act./Network & $\mathbf{1.30}$ & $0.92$ & & \\
& Act./Layer & $\mathbf{1.31}$ & $0.96$ & & \\ 

\bottomrule

\end{tabular}

\end{table}

\subsection{Lesion Experiments} \label{sec:lesion}

One approach that has been used for understanding both biological \citep{gazzaniga2013cognitive} and artificial \citep{zhou2018revisiting, casper2020frivolous} neural systems involves disrupting neurons during inference. 
We experiment with ``lesion'' tests in which we analyze network performance on the test set when a subcluster is dropped out.
We then analyze the damage to the network's performance.
First, we measure importance by taking the drop in accuracy.
Specifically, let $\theta$ be the parameter vector of the neural network, $c$ be a set of neurons, $\mathcal{M}(\theta, c)$ be a masked version of $\theta$ where weights into or out of nodes in $c$ have been set to 0, and $\textrm{Acc}(\vartheta, \mathcal{D})$ be the accuracy of the network parameterized by $\vartheta$ on dataset $\mathcal{D}$.
Then, our measure for importance is $\textrm{Acc}(\theta, \texttt{test}) - \textrm{Acc}(\mathcal{M}(\theta, c), \texttt{test})$, where $\texttt{test}$ is a test dataset that was not used to construct the activation-based partitionings.

Second, we measure the coherence in a subcluster with respect to class by taking the range of class-specific accuracy drops. 
Specifically, let $\texttt{test}_i$ be the subset of the test set with label $i$, and let $\Delta(\theta, c, i) := \textrm{Acc}(\theta, \texttt{test}_i) - \textrm{Acc}(\mathcal{M}(\theta, c), \texttt{test}_i)$ be the drop in accuracy for examples with label $i$ from lesioning $c$. 
Then, this measure of coherence is the range $\left(\max_i \Delta(\theta, c, i) - \min_i \Delta(\theta, c, i)\right)$, of accuracy drops over classes.
We use this to detect whether clusters are more crucial for some classes over others, which would suggest that they coherently act to correctly label those classes.

We use the analysis pipeline from Section \ref{sec:statistical_analysis} to test for importance and coherence using these overall accuracy differences and class-wise ranges.
In this setting, effect measures $>1$ indicate more importance/coherence among true subclusters on average compared to random ones. 
Table~\ref{tab:fisher_stats} shows Fisher statistics, and Table~\ref{tab:effect_sizes} shows effect measure data.
Results are summarized in Section~{\ref{subsec:findings}}. 

\begin{table}[h!]
\footnotesize
\caption{Effect measures for (1) lesion-based experiments measuring importance via overall accuracy drops (\textcolor{blue}{Acc. Drop}) and coherence via the class-wise range of accuracy drops (\textcolor{orange}{Class Range}); and (2) Feature visualization-based experiments in networks measuring coherence via the optimization score (\textcolor{orange}{Vis Score}) and the entropy of network outputs (\textcolor{orange}{Softmax $H$}). Each row corresponds to a network paired with a partitioning method. Results are calculated as explained in Section \ref{sec:statistical_analysis}.
For accuracy drop, class-wise range and visualization score experiments, an effect measure $>1$ corresponds to more importance/coherence among true subclusters than random ones, while one of $<1$ does for softmax entropy experiments.
Entries where the effect measure is more than two standard errors away from 1 in the direction of local specialization are \textbf{bolded}.}
\label{tab:effect_sizes}
\centerfloat
\begin{tabular}{l l r r r r }
\toprule
 & &  \multicolumn{2}{c}{\textbf{Lesion}}
& \multicolumn{2}{c}{\textbf{Feature Visualization}}\\

\textbf{Network} & \textbf{Partitioning}
& {\textbf{\textcolor{blue}{Acc.\ Drop}}}
& {\textbf{\textcolor{orange}{Class Range}}}
& {\textbf{\textcolor{orange}{Vis Score}}}
& {\textbf{\textcolor{orange}{Softmax $\mathbf{H}$}}} \\
& & {\textbf{\textcolor{blue}{High$\to$Imp.}}}
& {\textbf{\textcolor{orange}{High$\to$Coh.}}}
& {\textbf{\textcolor{orange}{High$\to$Coh.}}}
& {\textbf{\textcolor{orange}{Low$\to$Coh.}}}\\
\midrule

\multirow{4}{*}{MLP, MNIST} & Weight/Network & $\mathbf{1.123\pm0.058}$ & $0.701\pm0.036$ & ${1.003\pm0.004}$ & ${1.105\pm0.021}$ \\
& Weight/Layer & ${1.061\pm0.048}$ & $0.676\pm0.029$ & $\mathbf{1.024\pm0.004}$ & $\mathbf{0.931\pm0.016}$ \\
& Act./Network & ${0.883\pm0.038}$ & $0.646\pm0.024$ & $\mathbf{1.02\pm0.003}$ & ${0.997\pm0.013}$ \\
& Act./Layer & ${0.929\pm0.040}$ & $0.687\pm0.025$ & $\mathbf{1.026\pm0.003}$ & ${1.011\pm0.012}$ \\ \hline

\multirow{4}{*}{CNN, MNIST} & Weight/Network & ${0.837\pm0.048}$ & $0.527\pm0.031$ & $0.998\pm0.004$ & $1.026\pm0.008$ \\
& Weight/Layer & $0.814\pm0.046$ & $0.635\pm0.033$ & $1.004\pm0.003$ & $1.007\pm0.007$ \\
& Act./Network & ${1.078\pm0.061}$ & $0.543\pm0.039$ & $0.933\pm0.006$ & $1.025\pm0.011$  \\
& Act./Layer & ${0.939\pm0.060}$ & $0.625\pm0.043$ & $0.925\pm0.005$ & $\mathbf{0.970\pm0.010}$ \\ \hline

\multirow{4}{*}{VGG, CIFAR-10} & Weight/Network & ${0.682\pm0.066}$ & ${0.407\pm0.042}$ & ${0.871\pm0.011}$ & $1.124\pm0.012$ \\
& Weight/Layer & ${0.808\pm0.041}$ & ${0.692\pm0.033}$ & $\mathbf{1.013\pm0.006}$ & $0.992\pm0.012$ \\
& Act./Network & ${0.926\pm0.032}$ & $0.679\pm0.023$ & $\mathbf{1.327\pm0.005}$ & $\mathbf{0.950\pm0.009}$ \\
& Act./Layer & ${0.956\pm0.030}$ & $0.695\pm0.021$ & $\mathbf{1.379\pm0.004}$ & $\mathbf{0.930\pm0.008}$ \\ \hline

\multirow{4}{*}{VGG-16, ImageNet} & Weight/Network & $\mathbf{1.205\pm0.050}$ & $0.790\pm0.010$ & $\mathbf{1.043\pm0.005}$ & $0.998\pm0.001$ \\
& Weight/Layer & $\mathbf{1.168\pm0.019}$ & $0.825\pm0.005$ & $\mathbf{1.076\pm0.003}$ & $\mathbf{0.991\pm0.001}$ \\
& Act./Network & $\mathbf{1.129\pm0.026}$ & $0.859\pm0.006$ & $\mathbf{1.066\pm0.003}$ & $1.000\pm0.001$  \\
& Act./Layer & $\mathbf{1.063\pm0.021}$ & $0.876\pm0.005$ & $\mathbf{1.056\pm0.003}$ & $1.001\pm0.001$ \\ \hline

\multirow{4}{*}{VGG-19, ImageNet} & Weight/Network & & & $\mathbf{1.061\pm0.004}$ & $1.003\pm0.001$ \\
& Weight/Layer & & & $\mathbf{1.099\pm0.003}$ & $1.001\pm0.001$ \\
& Act./Network & & & $\mathbf{1.046\pm0.003}$ & $\mathbf{0.996\pm0.001}$ \\
& Act./Layer & & & $\mathbf{1.081\pm0.002}$ & $1.004\pm0.001$ \\ \hline

\multirow{4}{*}{ResNet18, ImageNet} & Weight/Network & ${0.926\pm0.045}$ & $0.957\pm0.011$ & & \\
& Weight/Layer & ${0.971\pm0.016}$ & $0.971\pm0.004$ & & \\
& Act./Network & ${0.979\pm0.017}$ & $0.977\pm0.004$ & & \\
& Act./Layer & ${0.983\pm0.014}$ & $0.967\pm0.004$ & & \\ 

\bottomrule

\end{tabular}

\end{table}

\subsection{Feature Visualization Experiments} \label{sec:feature_visualization}

To further analyze coherence, we leverage another set of interpretability techniques based on feature visualization. 
We use gradient-based optimization to create an input image which maximizes the $L_1$ norm of the pre-ReLU activations of the neurons in a subcluster. 
Letting the parameter vector be $\theta$ and the subcluster be $c$, we write $\textrm{Act}(x, \theta, c)$ for the vector of pre-ReLU activations of neurons in $c$ in network $\theta$ on input $x$, and denote this optimized input image as $x(\theta, c)$, which approximately maximizes $\|\textrm{Act}(x,\theta,c)\|_1$.
The key insight is that properties of these visualizations $x(\theta, c)$ can suggest what roles the subclusters play in the network.
Implementation details are in Appendix \ref{app:feature_visualization}, and Figure~\ref{fig:lucid_examples} shows example visualizations. 

We use two techniques to analyze coherence using these visualizations of subclusters. 
First, we analyzed the value of the maximization objective for each image we produced, $\|\textrm{Act}(x(\theta, c), \theta, c)\|_1$, which we call the ``score'' of the visualization. 
This gives one notion of how coherent a subcluster may be with respect to input features, because if a single image can strongly excite an entire subcluster, this suggests that the neurons comprising it are involved in detecting/processing related features.
Second, we obtain a measure of coherence by analyzing the entropy $H(\textrm{label} \mid x(\theta, c); \theta)$ of the softmax outputs of the network when these images are passed through. 
If the entropy is low, this suggests that a cluster is coherent with respect to class labels.

Just as with lesion experiments, we perform analysis using these two methods using the pipeline from Section \ref{sec:statistical_analysis} to measure how coherent true subclusters are compared to random ones. 

For visualization score experiments, effect measures $>1$ indicate coherence while for the softmax $H$ experiments, effect measures $<1$ indicate coherence. 
Table~\ref{tab:fisher_stats} shows Fisher statistics, and Table~\ref{tab:effect_sizes} shows effect measure data.

\subsection{Findings} \label{subsec:findings}

\textbf{Our partitionings identify important subclusters.} Fisher statistics for lesion accuracy drops are high and significant, as shown in Table~\ref{tab:fisher_stats}, indicating that sub-clusters are more likely to be highly important relative to random groups of neurons. However, as shown in Table~\ref{tab:effect_sizes}, not all of the corresponding effect measures are below one, even when the Fisher statistic is significantly greater than 0.98.
This indicates that when we detect that an unusual number of subclusters are important, this does not necessarily correspond to importance on average. 

\textbf{Our partitionings identify subclusters that are coherent w.r.t. input features but not class label.} Class-specific measures of coherence, class-wise lesion accuracy drop range and output entropy, showed significant coherence in almost no conditions. The class-wise range measure even tended to show that subclusters were less coherent w.r.t. class than random groups of neurons.
However, subclusters were reliably coherent as measured by \emph{visualization score}, both as quantified by Fisher statistics and effect measures.
Together, these results offer evidence that subclusters tended to perform coherent sub-tasks, but not in a class-specific way. 

\textbf{All partitioning methods yield similar results.} In Table~\ref{tab:fisher_stats}, we find no clear difference between the Fisher statistics of activation-based and weight-based clusterings, or between layer-wise and network-wide clusterings.
This is somewhat unexpected:
one might have predicted that weight-based methods' lack of runtime information or layer-wise methods' lack of global information would lead to lower quality clusterings, but this was not the case.

\section{Discussion} \label{sec:discussion}

\textbf{Contributions:} In this work, we introduce several methods for partitioning networks into clusters of neurons and analyzing the resulting partitions for local specialization.
Key to this is measuring proxies: \emph{importance} as a means of understanding what parts of a network are crucial for performance, and \emph{coherence} as a measure for how much the neurons in a part work together.
We rigorously evaluate these proxies using statistical methods, finding that even the weights-only clustering methods
are able to reveal clusters with a significant degree of importance and coherence compared to random ones. 
In each network, we found evidence that our partitioning methods were able to identify
specialized subsets of neurons
via measuring accuracy drops under lesions (importance) and feature visualization scores (coherence).
To the best of our knowledge, ours is the first method which is able to quantitatively assess the local specialization of neural networks in a way that does not require a human in the loop.

\textbf{Relation to other research:} Having effective tools for interpreting networks is important for understanding AI systems, in particular by helping to diagnose failure modes (e.g., \citet{carter2019activation, mu2020compositional, casper2021one}).
Our work relates to this goal, though indirectly.
The tests we perform are based on data from lesions and feature visualizations, both of which are interpretability tools. 
But rather than directly using these data to interpret subclusters, our focus is one step higher: on automatedly testing whether these subclusters are worth analyzing at all and finding ways to screen for subclusters that should be the subject of deeper investigation. 
By showing that the partitioning methods we use 
generate partitionings that align with local specialization,
these results suggest that clustering neurons offers a useful level of abstraction through which to study networks.

\textbf{Limitations:} One limitation of our work is a lack of assurance that importance and coherence are reliably strong proxies for human-comprehensible forms of local specialization.
While they are sufficient to imply some degree of abstractability with respect to the task at hand, they may not always be particularly useful for understanding the network.
Relatedly, our approach is also not designed to identify the sub-task performed by a group of neurons, nor does it identify relationships between groups. 
Given these limitations, the tools we introduce should be seen as methods for screening a network for evidence of local specialization overall and for particular sets of neurons where sub-task functionality is localized.
A final notable limitation is that these methods do not offer tools for building more modular networks beyond techniques for measuring local specialization. 
Future work toward this may benefit from our techniques but should also hinge on architectural or regularization-based methods for promoting independent operation of groups of neurons.

\textbf{Conclusion:} While we make progress here toward better understanding how networks can be understood, neural systems are still complex, and more insights are needed to develop useful understandings of them.
The ultimate goal should be to develop reliable methods for building models which perform well and also lend themselves to faithful abstract interpretations.
We believe that these methods should involve testing networks for local specialization and investigating the functions that important or coherent parts of the network specialize in.

\section*{Acknowledgments}
The authors would like to thank Open Philanthropy for funding to support this project, the researchers at UC
Berkeley’s Center for Human-Compatible AI for their advice, and anonymous reviewers for their contributions to improving the paper.

\small
\bibliographystyle{iclr2021_conference}
\bibliography{references}

\begin{thebibliography}{45}
\providecommand{\natexlab}[1]{#1}
\providecommand{\url}[1]{\texttt{#1}}
\expandafter\ifx\csname urlstyle\endcsname\relax
  \providecommand{\doi}[1]{doi: #1}\else
  \providecommand{\doi}{doi: \begingroup \urlstyle{rm}\Url}\fi

\bibitem[Abadi et~al.(2015)Abadi, Agarwal, Barham, Brevdo, Chen, Citro,
  Corrado, Davis, Dean, Devin, Ghemawat, Goodfellow, Harp, Irving, Isard, Jia,
  Jozefowicz, Kaiser, Kudlur, Levenberg, Man\'{e}, Monga, Moore, Murray, Olah,
  Schuster, Shlens, Steiner, Sutskever, Talwar, Tucker, Vanhoucke, Vasudevan,
  Vi\'{e}gas, Vinyals, Warden, Wattenberg, Wicke, Yu, and
  Zheng]{tensorflow2015-whitepaper}
Mart\'{\i}n Abadi, Ashish Agarwal, Paul Barham, Eugene Brevdo, Zhifeng Chen,
  Craig Citro, Greg~S. Corrado, Andy Davis, Jeffrey Dean, Matthieu Devin,
  Sanjay Ghemawat, Ian Goodfellow, Andrew Harp, Geoffrey Irving, Michael Isard,
  Yangqing Jia, Rafal Jozefowicz, Lukasz Kaiser, Manjunath Kudlur, Josh
  Levenberg, Dandelion Man\'{e}, Rajat Monga, Sherry Moore, Derek Murray, Chris
  Olah, Mike Schuster, Jonathon Shlens, Benoit Steiner, Ilya Sutskever, Kunal
  Talwar, Paul Tucker, Vincent Vanhoucke, Vijay Vasudevan, Fernanda Vi\'{e}gas,
  Oriol Vinyals, Pete Warden, Martin Wattenberg, Martin Wicke, Yuan Yu, and
  Xiaoqiang Zheng.
\newblock {TensorFlow}: Large-scale machine learning on heterogeneous systems,
  2015.
\newblock URL \url{https://www.tensorflow.org/}.
\newblock Software available from tensorflow.org.

\bibitem[Alet et~al.(2018)Alet, Lozano-P{\'e}rez, and
  Kaelbling]{alet2018modular}
Ferran Alet, Tom{\'a}s Lozano-P{\'e}rez, and Leslie~P Kaelbling.
\newblock Modular meta-learning.
\newblock \emph{arXiv preprint arXiv:1806.10166}, 2018.

\bibitem[Baldwin \& Clark(2000)Baldwin and Clark]{baldwin2000design}
Carliss~Young Baldwin and Kim~B Clark.
\newblock \emph{Design rules: The power of modularity}, volume~1.
\newblock MIT press, 2000.

\bibitem[Bau et~al.(2017)Bau, Zhou, Khosla, Oliva, and
  Torralba]{bau2017network}
David Bau, Bolei Zhou, Aditya Khosla, Aude Oliva, and Antonio Torralba.
\newblock Network dissection: Quantifying interpretability of deep visual
  representations.
\newblock In \emph{Proceedings of the IEEE conference on computer vision and
  pattern recognition}, pp.\  6541--6549, 2017.

\bibitem[Benjamini \& Hochberg(1995)Benjamini and
  Hochberg]{benjamini1995controlling}
Yoav Benjamini and Yosef Hochberg.
\newblock Controlling the false discovery rate: a practical and powerful
  approach to multiple testing.
\newblock \emph{Journal of the Royal statistical society: series B
  (Methodological)}, 57\penalty0 (1):\penalty0 289--300, 1995.

\bibitem[Booch et~al.(2007)Booch, Maksimchuk, Engle, Young, Conallen, and
  Houston]{booch2007object}
Grady Booch, Robert~A Maksimchuk, Michael~W Engle, Bobbi Young, Jim Conallen,
  and Kelli~A Houston.
\newblock \emph{Object-Oriented Analysis and Design with Applications}.
\newblock Addison-Wesley Professional, third edition, 2007.

\bibitem[Borz{\`{i}} \& Borz{\`{i}}(2006)Borz{\`{i}} and
  Borz{\`{i}}]{borzi2006algebraic}
Alfio Borz{\`{i}} and Giuseppe Borz{\`{i}}.
\newblock Algebraic multigrid methods for solving generalized eigenvalue
  problems.
\newblock \emph{International journal for numerical methods in engineering},
  65\penalty0 (8):\penalty0 1186--1196, 2006.

\bibitem[Cammarata et~al.(2020)Cammarata, Carter, Goh, Olah, Petrov, and
  Schubert]{cammarata2020thread}
Nick Cammarata, Shan Carter, Gabriel Goh, Chris Olah, Michael Petrov, and
  Ludwig Schubert.
\newblock Thread: Circuits.
\newblock \emph{Distill}, 2020.
\newblock \doi{10.23915/distill.00024}.
\newblock \url{https://distill.pub/2020/circuits}.

\bibitem[Carter et~al.(2019)Carter, Armstrong, Schubert, Johnson, and
  Olah]{carter2019activation}
Shan Carter, Zan Armstrong, Ludwig Schubert, Ian Johnson, and Chris Olah.
\newblock Activation atlas.
\newblock \emph{Distill}, 4\penalty0 (3):\penalty0 e15, 2019.

\bibitem[Casper et~al.(2020)Casper, Boix, D'Amario, Guo, Vinken, and
  Kreiman]{casper2020frivolous}
Stephen Casper, Xavier Boix, Vanessa D'Amario, Ling Guo, Kasper Vinken, and
  Gabriel Kreiman.
\newblock Frivolous units: Wider networks are not really that wide.
\newblock \emph{arXiv preprint arXiv:1912.04783}, 2020.

\bibitem[Casper et~al.(2021)Casper, Nadeau, and Kreiman]{casper2021one}
Stephen Casper, Max Nadeau, and Gabriel Kreiman.
\newblock One thing to fool them all: Generating interpretable, universal, and
  physically-realizable adversarial features.
\newblock \emph{arXiv preprint arXiv:2110.03605}, 2021.

\bibitem[Chollet et~al.(2015)]{chollet2015keras}
Fran\c{c}ois Chollet et~al.
\newblock Keras.
\newblock \url{https://keras.io}, 2015.

\bibitem[Clune et~al.(2013)Clune, Mouret, and Lipson]{clune2013evolutionary}
Jeff Clune, Jean-Baptiste Mouret, and Hod Lipson.
\newblock The evolutionary origins of modularity.
\newblock \emph{Proceedings of the Royal Society {B}: Biological sciences},
  280\penalty0 (1755), 2013.

\bibitem[Csord{\'a}s et~al.(2021)Csord{\'a}s, van Steenkiste, and
  Schmidhuber]{csordasneural2020}
R{\'o}bert Csord{\'a}s, Sjoerd van Steenkiste, and J{\"u}rgen Schmidhuber.
\newblock Are neural nets modular? inspecting their functionality through
  differentiable weight masks.
\newblock In \emph{International Conference on Learning Representations}, 2021.

\bibitem[De~Lange et~al.(2019)De~Lange, Aljundi, Masana, Parisot, Jia,
  Leonardis, Slabaugh, and Tuytelaars]{de2019continual}
Matthias De~Lange, Rahaf Aljundi, Marc Masana, Sarah Parisot, Xu~Jia,
  Ale{\v{s}} Leonardis, Gregory Slabaugh, and Tinne Tuytelaars.
\newblock A continual learning survey: Defying forgetting in classification
  tasks.
\newblock \emph{arXiv preprint arXiv:1909.08383}, 2019.

\bibitem[Filan et~al.(2021)Filan, Casper, Hod, Wild, Critch, and
  Russell]{filan2021clusterability}
Daniel Filan, Stephen Casper, Shlomi Hod, Cody Wild, Andrew Critch, and Stuart
  Russell.
\newblock Clusterability in neural networks.
\newblock \emph{arXiv preprint arXiv:2103.03386}, 2021.

\bibitem[Gazzaniga \& Ivry(2013)Gazzaniga and Ivry]{gazzaniga2013cognitive}
Michael Gazzaniga and Richard~B Ivry.
\newblock \emph{Cognitive Neuroscience: The Biology of the Mind: Fourth
  International Student Edition}.
\newblock WW Norton, 2013.

\bibitem[Girvan \& Newman(2002)Girvan and Newman]{girvan2002community}
Michelle Girvan and Mark~EJ Newman.
\newblock Community structure in social and biological networks.
\newblock \emph{Proceedings of the national academy of sciences}, 99\penalty0
  (12):\penalty0 7821--7826, 2002.

\bibitem[Goyal et~al.(2019)Goyal, Lamb, Hoffmann, Sodhani, Levine, Bengio, and
  Sch{\"o}lkopf]{goyal2019recurrent}
Anirudh Goyal, Alex Lamb, Jordan Hoffmann, Shagun Sodhani, Sergey Levine,
  Yoshua Bengio, and Bernhard Sch{\"o}lkopf.
\newblock Recurrent independent mechanisms.
\newblock \emph{arXiv preprint arXiv:1909.10893}, 2019.

\bibitem[He et~al.(2016)He, Zhang, Ren, and Sun]{he2016deep}
Kaiming He, Xiangyu Zhang, Shaoqing Ren, and Jian Sun.
\newblock Deep residual learning for image recognition.
\newblock In \emph{Proceedings of the IEEE conference on computer vision and
  pattern recognition}, pp.\  770--778, 2016.

\bibitem[Kingma \& Ba(2014)Kingma and Ba]{kingma2014adam}
Diederik~P Kingma and Jimmy Ba.
\newblock Adam: A method for stochastic optimization.
\newblock \emph{arXiv preprint arXiv:1412.6980}, 2014.

\bibitem[Kirsch et~al.(2018)Kirsch, Kunze, and Barber]{kirsch2018modular}
Louis Kirsch, Julius Kunze, and David Barber.
\newblock Modular networks: Learning to decompose neural computation.
\newblock In \emph{Advances in Neural Information Processing Systems}, pp.\
  2408--2418, 2018.

\bibitem[Krizhevsky \& Hinton(2009)Krizhevsky and
  Hinton]{krizhevsky2009learning}
Alex Krizhevsky and Geoffrey Hinton.
\newblock Learning multiple layers of features from tiny images.
\newblock Technical report, University of Toronto, 2009.

\bibitem[Lake et~al.(2015)Lake, Salakhutdinov, and Tenenbaum]{lake2015human}
Brenden~M Lake, Ruslan Salakhutdinov, and Joshua~B Tenenbaum.
\newblock Human-level concept learning through probabilistic program induction.
\newblock \emph{Science}, 350\penalty0 (6266):\penalty0 1332--1338, 2015.

\bibitem[Lake et~al.(2017)Lake, Ullman, Tenenbaum, and
  Gershman]{lake2017building}
Brenden~M Lake, Tomer~D Ullman, Joshua~B Tenenbaum, and Samuel~J Gershman.
\newblock Building machines that learn and think like people.
\newblock \emph{Behavioral and brain sciences}, 40, 2017.

\bibitem[LeCun et~al.(1998)LeCun, Bottou, Bengio, and
  Haffner]{lecun1998gradient}
Yann LeCun, L{\'e}on Bottou, Yoshua Bengio, and Patrick Haffner.
\newblock Gradient-based learning applied to document recognition.
\newblock \emph{Proceedings of the IEEE}, 86\penalty0 (11):\penalty0
  2278--2324, 1998.

\bibitem[Liu \& Deng(2015)Liu and Deng]{DBLP:conf/acpr/LiuD15}
Shuying Liu and Weihong Deng.
\newblock Very deep convolutional neural network based image classification
  using small training sample size.
\newblock In \emph{3rd {IAPR} Asian Conference on Pattern Recognition, {ACPR}
  2015, Kuala Lumpur, Malaysia, November 3-6, 2015}, pp.\  730--734. {IEEE},
  2015.
\newblock \doi{10.1109/ACPR.2015.7486599}.
\newblock URL \url{https://doi.org/10.1109/ACPR.2015.7486599}.

\bibitem[Madan et~al.(2020)Madan, Henry, Dozier, Ho, Bhandari, Sasaki, Durand,
  Pfister, and Boix]{madan2020capability}
Spandan Madan, Timothy Henry, Jamell Dozier, Helen Ho, Nishchal Bhandari,
  Tomotake Sasaki, Fr{\'e}do Durand, Hanspeter Pfister, and Xavier Boix.
\newblock On the capability of neural networks to generalize to unseen
  category-pose combinations.
\newblock \emph{arXiv preprint arXiv:2007.08032}, 2020.

\bibitem[Marengo et~al.(2017)Marengo, Farnsworth, and
  Stefanic]{marengo2017geometric}
James~E Marengo, David~L Farnsworth, and Lucas Stefanic.
\newblock A geometric derivation of the irwin-hall distribution.
\newblock \emph{International Journal of Mathematics and Mathematical
  Sciences}, 2017, 2017.

\bibitem[Morcos et~al.(2018)Morcos, Barrett, Rabinowitz, and
  Botvinick]{morcos2018importance}
Ari~S Morcos, David~GT Barrett, Neil~C Rabinowitz, and Matthew Botvinick.
\newblock On the importance of single directions for generalization.
\newblock \emph{arXiv preprint arXiv:1803.06959}, 2018.

\bibitem[Mu \& Andreas(2020)Mu and Andreas]{mu2020compositional}
Jesse Mu and Jacob Andreas.
\newblock Compositional explanations of neurons.
\newblock \emph{arXiv preprint arXiv:2006.14032}, 2020.

\bibitem[Newman \& Girvan(2004)Newman and Girvan]{newman2004finding}
Mark~EJ Newman and Michelle Girvan.
\newblock Finding and evaluating community structure in networks.
\newblock \emph{Physical review E}, 69\penalty0 (2):\penalty0 026113, 2004.

\bibitem[Olah et~al.(2017)Olah, Mordvintsev, and Schubert]{olah2017feature}
Chris Olah, Alexander Mordvintsev, and Ludwig Schubert.
\newblock Feature visualization.
\newblock \emph{Distill}, 2\penalty0 (11):\penalty0 e7, 2017.

\bibitem[Panzeri et~al.(2017)Panzeri, Harvey, Piasini, Latham, and
  Fellin]{panzeri2017cracking}
Stefano Panzeri, Christopher~D Harvey, Eugenio Piasini, Peter~E Latham, and
  Tommaso Fellin.
\newblock Cracking the neural code for sensory perception by combining
  statistics, intervention, and behavior.
\newblock \emph{Neuron}, 93\penalty0 (3):\penalty0 491--507, 2017.

\bibitem[Parascandolo et~al.(2018)Parascandolo, Kilbertus, Rojas-Carulla, and
  Sch{\"o}lkopf]{parascandolo2018learning}
Giambattista Parascandolo, Niki Kilbertus, Mateo Rojas-Carulla, and Bernhard
  Sch{\"o}lkopf.
\newblock Learning independent causal mechanisms.
\newblock In \emph{International Conference on Machine Learning}, pp.\
  4036--4044. PMLR, 2018.

\bibitem[Pedregosa et~al.(2011)Pedregosa, Varoquaux, Gramfort, Michel, Thirion,
  Grisel, Blondel, Prettenhofer, Weiss, Dubourg, et~al.]{pedregosa2011scikit}
Fabian Pedregosa, Ga{\"e}l Varoquaux, Alexandre Gramfort, Vincent Michel,
  Bertrand Thirion, Olivier Grisel, Mathieu Blondel, Peter Prettenhofer, Ron
  Weiss, Vincent Dubourg, et~al.
\newblock Scikit-learn: Machine learning in python.
\newblock \emph{the Journal of machine Learning research}, 12:\penalty0
  2825--2830, 2011.

\bibitem[Shi \& Malik(2000)Shi and Malik]{shi2000normalized}
Jianbo Shi and Jitendra Malik.
\newblock Normalized cuts and image segmentation.
\newblock \emph{{IEEE} Transactions on pattern analysis and machine
  intelligence}, 22\penalty0 (8):\penalty0 888--905, 2000.

\bibitem[Simonyan \& Zisserman(2014)Simonyan and Zisserman]{simonyan2014very}
Karen Simonyan and Andrew Zisserman.
\newblock Very deep convolutional networks for large-scale image recognition.
\newblock \emph{arXiv preprint arXiv:1409.1556}, 2014.

\bibitem[Testolin et~al.(2020)Testolin, Piccolini, and
  Suweis]{testolin2020deep}
Alberto Testolin, Michele Piccolini, and Samir Suweis.
\newblock Deep learning systems as complex networks.
\newblock \emph{Journal of Complex Networks}, 8\penalty0 (1):\penalty0 cnz018,
  2020.

\bibitem[Udrescu et~al.(2020)Udrescu, Tan, Feng, Neto, Wu, and
  Tegmark]{udrescu2020ai}
Silviu-Marian Udrescu, Andrew Tan, Jiahai Feng, Orisvaldo Neto, Tailin Wu, and
  Max Tegmark.
\newblock Ai feynman 2.0: Pareto-optimal symbolic regression exploiting graph
  modularity.
\newblock \emph{arXiv preprint arXiv:2006.10782}, 2020.

\bibitem[{v}on Luxburg(2007)]{von2007tutorial}
Ulrike {v}on Luxburg.
\newblock A tutorial on spectral clustering.
\newblock \emph{Statistics and computing}, 17\penalty0 (4):\penalty0 395--416,
  2007.

\bibitem[Voss et~al.(2021)Voss, Goh, Cammarata, Petrov, Schubert, and
  Olah]{voss2021branch}
Chelsea Voss, Gabriel Goh, Nick Cammarata, Michael Petrov, Ludwig Schubert, and
  Chris Olah.
\newblock Branch specialization.
\newblock \emph{Distill}, 6\penalty0 (4):\penalty0 e00024--008, 2021.

\bibitem[Watanabe(2019)]{watanabe2019interpreting}
Chihiro Watanabe.
\newblock Interpreting layered neural networks via hierarchical modular
  representation.
\newblock In \emph{International Conference on Neural Information Processing},
  pp.\  376--388. Springer, 2019.

\bibitem[You et~al.(2020)You, Leskovec, He, and Xie]{you2020graph}
Jiaxuan You, Jure Leskovec, Kaiming He, and Saining Xie.
\newblock Graph structure of neural networks.
\newblock \emph{arXiv preprint arXiv:2007.06559}, 2020.

\bibitem[Zhou et~al.(2018)Zhou, Sun, Bau, and Torralba]{zhou2018revisiting}
Bolei Zhou, Yiyou Sun, David Bau, and Antonio Torralba.
\newblock Revisiting the importance of individual units in {CNNs} via ablation.
\newblock \emph{arXiv preprint arXiv:1806.02891}, 2018.

\end{thebibliography}
\normalsize
\newpage

\appendix

\section{Appendix}

\subsection{Spectral Clustering Algorithm}
\label{app:spectral_clustering_algorithm}

The spectral clustering algorithm on the graph $G = (V,E)$ produces a partition of its vertices in which there are stronger connections within sets of vertices than between them \citep{shi2000normalized}. 
It does so by solving a relaxation of the NP-Hard problem of minimizing the \emph{n-cut} (normalized cut) for a partition. 
For disjoint, non-empty sets $X_1, ... X_k$ where $\cup_{i=1}^k X_i = V$, this is defined by \citet{von2007tutorial} as:
$$\operatorname{n-cut}(X_1, ..., X_k)
\coloneqq \frac{1}{2} \sum_{i=1}^{k} \frac{W(X_i, \overline{X_i})}{\mathrm{vol}(X_i)}$$
for two sets of vertices $X, Y \subseteq V$, we define $W(X, Y) \coloneqq \sum_{v_i \in X, v_j \in Y} w_{ij}$;
the degree of a vertex $v_i \in V$ is $d_i = \sum_{j=1}^{n} w_{ij}$;
and the volume of a subset $X \subseteq V$ is $\mathrm{vol}(X) \coloneqq \sum_{i \in X} d_i$. 

We use the \emph{scikit-learn} implementation \citep{pedregosa2011scikit} with the ARPACK eigenvalue solver \citep{borzi2006algebraic}. 

\begin{algorithm}[h!]
\SetKwInOut{Input}{Input}
\SetKwInOut{Output}{Output}
\SetAlgoLined
\LinesNumbered
\Input{Weighted adjacency matrix $W \in \mathbb{R}^{n \times n}$, number $k$ of clusters to construct}
 
 Compute the unnormalized Laplacian $L$.
 
 Compute the first $k$ generalized eigenvectors $u_1, ..., u_k$ of the generalized eigenproblem $L u = \lambda D u$.
 
 Let $U \in \mathbb{R}^{n \times k}$ be the matrix containing the vectors $u_1, ..., u_k$ as columns.
 
 For $i = 1, .., n$, let $y_i \in \mathbb{R}^k$ be the vector corresponding to the $i$\textsuperscript{th} row of $U$.
 
 Cluster the points $(y_i)_{i=1,...,n}$ in $\mathbb{R}^k$ with the \emph{k-means} algorithm into clusters $C_1, ..., C_k$,
 
 \Output{Clusters $A_1, ..., A_k$ with $A_i = \{j| y_j \in C_i\}$.}
 
 \caption{\textbf{Normalized spectral clustering} according to \citet{shi2000normalized}, implemented in \emph{scikit-learn} \citep{pedregosa2011scikit}, description taken from \citet{von2007tutorial}.}
 \label{alg:spectral}
\end{algorithm}

\subsection{Network Training Details}
\label{app:network_training_details}

We use Tensorflow's implementation of the Keras API \citep{tensorflow2015-whitepaper, chollet2015keras}. When training all networks, we use the Adam algorithm \citep{kingma2014adam} with the standard Keras hyperparameters: learning rate $0.001$, $\beta_1 = 0.9$, $\beta_2 = 0.999$, no amsgrad. 
The loss function was categorical cross-entropy. 
For all non-ImageNet networks, we train 5 identically-configured replicates. 

\textbf{Small MLPs (MNIST):} 
We train MLPs with 4 hidden layers, each of width 256, for 20 epochs with batch size 128. 
All MLPs achieved a test accuracy on MNIST between 97.6\% and 98.2\%. 
On Fashion-MNIST, they achieved test accuracies between 88.4\% and 89.4\%.

\textbf{Small CNNs (MNIST):} 
These networks had 3 convolutional layers with 64 $3 \times 3$ channels each with the second and third hidden layers being followed by max pooling with a 2 by 2 window. 
There was a final fully-connected hidden layer with 128 neurons. 
We train them with a batch size of 64 for 10 epochs. 
All small CNNs achieved a testing accuracy MNIST between 99.1\% and 99.4\%.
On Fashion-MNIST, they achieved test accuracies between 91.8\% and 92.4\%.

\textbf{Mid-sized VGG CNNs (CIFAR-10):} 
We implement a version of VGG-16 described by \citet{simonyan2014very, DBLP:conf/acpr/LiuD15}. 
We train these with Adam, for 200 epochs with a batch size of 128. 
These are trained using $L_2$ regularization with a coefficient of $5 \times 10^{-4}$ and dropout with a rate tuned per-layer as done in \citet{simonyan2014very, DBLP:conf/acpr/LiuD15}.
Training was done with data augmentation which consisted of random rotations between 0 and 15 degrees, random shifts both vertically and horizontally of up to 10\% of the side length, and random horizontal flipping.
Test accuracies were between 82.8\% and 86.6\%.

\textbf{Large CNNs (ImageNet):} We experimented with VGG-16 and 19 \citep{simonyan2014very} and ResNet-18, and 50 \citep{he2016deep} networks. Weights were obtained from the Python \texttt{image-classifiers} package, version 1.0.0.

\newpage

\subsection{Pipeline - Second Part} \label{app:pipeline-second-part}

\begin{figure}[h!]
    \centering
    \caption{\textbf{Our extended procedural pipeline.} This figure expands Figure \ref{fig:schematic} and shows the successive steps after generating a partitioning of subclusters (step 3 in Figure \ref{fig:schematic}). After performing either lesion or feature visualization analysis, the results from each true subcluster and its random subclusters are aggregated to produce $p$ values and effect measures. For simplicity, only the analysis for the lesion experiment is presented, but the same pipeline is used for the feature visualization experiments.}
    \label{fig:schematic-extension}
    \includegraphics[width=1.1\columnwidth]{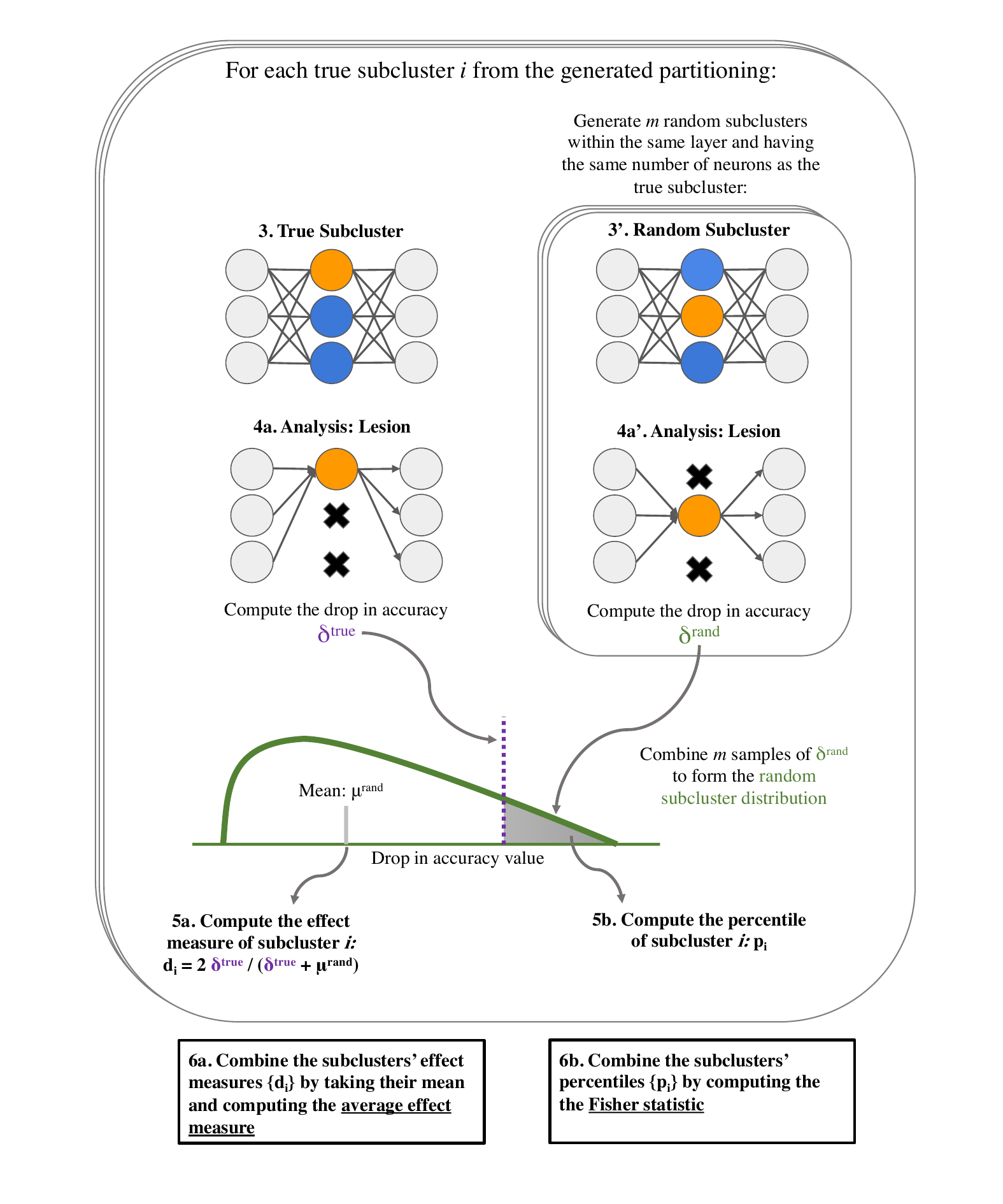}
    
\end{figure}

\newpage

\subsection{On Redundancy} \label{app:on_redundancy}

If multiple neurons in a network are redundant, having similar weights and activations, our clustering methods will likely group them. These groups of redundant neurons would be coherent and perhaps important, thereby demonstrating functional locality in a way that is arguably less interesting than for a non-atomic subtask. However, two pieces of evidence suggest that redundancy is not the sole driver of our results.

First, activation-based clusterings that group neurons with correlated activity are more likely to detect redundancy than weight-based clusterings. However, Table~\ref{tab:fisher_stats} shows that the activation-based clusterings are not reliably more important or coherent than weight-based clusterings, suggesting that the importance and coherence is not only due to redundancy.

Second, in addition to the VGGs trained on CIFAR-10 with L2 regularization and dropout that we analyze in Table~\ref{tab:fisher_stats}, we also train and analyze unregularized versions. Since dropout encourages redundant information to be encoded in multiple neurons, and L2 regularization encourages each neuron to depend on many previous-layer neurons, these unregularized networks should have less redundancy than those analyzed in the main paper. Our results in Table~\ref{tab:lesion_unreg} show unregularized networks as having more importance and coherence for many networks and metrics, suggesting that our measures are not just picking up redundancy.

\begin{table}[h!]
\footnotesize
\caption{Comparison for regularized and unregularized CNN-VGG networks for lesion-based experiments. Results are shown for $k \in \{8, 12, 16\}$. Fisher statistics are means over 5 runs. Each Fisher statistic which is significant at the $\alpha=0.05$ after Benjamini-Hochberg correction is \textbf{bold}. An effect measure $>1$ corresponds to more importance/coherence among true subclusters than random ones. All above 1 are \textbf{bold}.}
\label{tab:lesion_unreg}
\begin{center}
\begin{tabular}{l l r r r r }

\toprule
 & &  \multicolumn{2}{c}{\textcolor{blue}{\textbf{Importance: Acc. Drop}}} & \multicolumn{2}{c}{\textcolor{orange}{\textbf{Coherence: Class Range}}}\\
\textbf{Network} & \textbf{Partitioning} & {\textbf{Fisher}} & {\textbf{Effect Meas.}}  & {\textbf{Fisher}} & {\textbf{Effect Meas.}}  \\
& & \textbf{stat.} & \textbf{High$\to$Imp.} & \textbf{stat.} & \textbf{High$\to$Coh.}  \\

\midrule

& Weight/Network & $\mathbf{1.50}$ & $0.682 \pm 0.066$ & $\mathbf{2.12}$ & $0.407 \pm 0.042$ \\
VGG, CIFAR-10 & Weight/Layer & $\mathbf{1.15}$ & $0.808 \pm 0.041$ & $\mathbf{1.27}$ & $0.692 \pm 0.033$ \\
Regularized & Act./Network & $\mathbf{1.40}$ & $0.926 \pm 0.032$ & $0.97$ & $0.679 \pm 0.023$ \\
& Act./Layer & $\mathbf{1.56}$ & $0.956 \pm 0.030$ & $1.03$ & $0.695 \pm 0.021$ \\
\hline

& Weight/Network & $\mathbf{1.27}$ & $\mathbf{1.033 \pm 0.007}$ & $1.13$ & $0.995 \pm 0.015$ \\
VGG, CIFAR-10 & Weight/Layer & $0.94$ & $0.994 \pm 0.004$ & $0.92$ & $\mathbf{1.042 \pm 0.011}$ \\
Unregularized & Act./Network & $\mathbf{1.48}$ & $\mathbf{1.032 \pm 0.003}$ & $\mathbf{1.25}$ & $\mathbf{1.022 \pm 0.011}$ \\
& Act./Layer & $\mathbf{1.76}$ & $\mathbf{1.049 \pm 0.003}$ & $\mathbf{1.17}$ & $\mathbf{1.077 \pm 0.010}$ \\

\bottomrule
\end{tabular}
\end{center}
\end{table}

\begin{table}[h!]
\footnotesize
\caption{Comparison for regularized and unregularized CNN-VGG networks for feature visualization-based experiments. Results are shown for $k \in \{8,12,16\}$. Fisher statistics are means over 5 runs. Each mean Fisher statistic which is significant at the $\alpha=0.05$ level after Benjamini-Hochberg correction is \textbf{bold}. For vis score tests, an effect measure $>1$ corresponds to more coherence, and for softmax $H$ tests, one of $<1$ corresponds to more coherence. All effect measures on the side of 1 indicating more coherence are \textbf{bold}.}
\label{tab:lucid_unreg}
\begin{center}
\begin{tabular}{l l r r r r}
\toprule
 & &  \multicolumn{2}{c}{\textcolor{orange}{\textbf{Coherence: Vis Score}}} & \multicolumn{2}{c}{\textcolor{orange}{\textbf{Coherence: Softmax $\mathbf{H}$}}}\\
\textbf{Network} & \textbf{Partitioning} & {\textbf{Fisher}} & {\textbf{Effect Meas.}}  & {\textbf{Fisher}} & {\textbf{Effect Meas.}}  \\
& & \textbf{stat.} & \textbf{High$\to$Coh.} & \textbf{stat.} & \textbf{Low$\to$Coh.}  \\
\midrule

& Weight/Network & $\mathbf{1.46}$ & $0.871 \pm 0.011$ & $0.99$ & $1.124 \pm 0.012$ \\
VGG, CIFAR-10 & Weight/Layer & $1.00$ & $\mathbf{1.013 \pm 0.006}$ & $0.97$ & $0.992 \pm 0.012$ \\
Regularized & Act./Network & $\mathbf{2.34}$ & $\mathbf{1.327 \pm 0.005}$ & $1.08$ & $\mathbf{0.950 \pm 0.009}$ \\
& Act./Layer & $\mathbf{2.67}$ & $\mathbf{1.379 \pm 0.004}$ & $\mathbf{1.12}$ & $\mathbf{0.930 \pm 0.008}$ \\ \hline

& Weight/Network & $\mathbf{1.27}$ & $\mathbf{1.033 \pm 0.007}$ & $1.13$ & $0.995 \pm 0.015$ \\
VGG, CIFAR-10 & Weight/Layer & $0.94$ & $0.994 \pm 0.004$ & $0.92$ & $1.0424 \pm 0.011$ \\
Unregularized & Act./Network & $\mathbf{1.48}$ & $\mathbf{1.032 \pm 0.003}$ & $\mathbf{1.25}$ & $1.022 \pm 0.011$ \\
& Act./Layer & $\mathbf{1.76}$ & $\mathbf{1.049 \pm 0.003}$ & $\mathbf{1.17}$ & $1.077 \pm 0.010$ \\

\bottomrule
\end{tabular}
\end{center}
\end{table}

\subsection{Correlation-Based Visualization} \label{app:correlation_based_visualization}

\begin{figure}[t!]
    \centering
    \includegraphics[width=0.93\columnwidth]{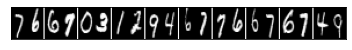}
    \includegraphics[width=0.9\columnwidth]{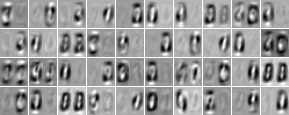}
    \caption{\textbf{An illustrative example of coherence.} Randomly selected examples of class ``3'' from our ``halves-MNIST'' dataset (top) and visualizations of neuron clusters in the first layer of networks trained to output the modular sum of digits in the images (bottom). }
    \label{fig:halves_vis}
\end{figure}

Figure \ref{fig:halves_vis} shows an illustrative example of coherence. We trained multilayer perceptrons on a version of the MNIST dataset \citep{lecun1998gradient} in which the images were two half-width digits side-by-side and the labels were their sum modulo 10. We clustered their first-layer neurons using approaches presented in Section \ref{sec:partition}. We then used a correlation-based method from \citet{watanabe2019interpreting} to create visualizations of the clusters. Almost all show selectivity to one half of the input. Here, we detail our approach and compare to visualizations of random sets of units.

\begin{figure*}[h!]
\centering
\begin{tabular}{lc}
    (a) & \includegraphics[width=0.4\columnwidth]{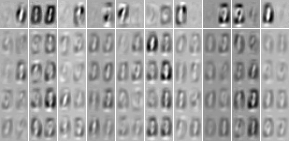}\\
    (b) & \includegraphics[width=0.32\columnwidth]{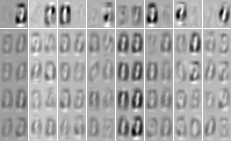}\\
    (c) & \includegraphics[width=0.4\columnwidth]{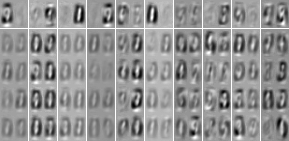}\\
    (d) & \includegraphics[width=0.56\columnwidth]{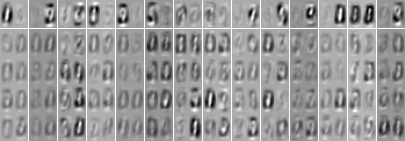} 
\end{tabular}
\caption{Comparison of true and random subcluster visualizations for the first layer of MLPs trained on the halves-MNIST dataset. The first rows show visualizations for true subclusters, and the bottom four show visualizations for random ones of the same size. Each panel gives results for a different approach to clustering: (a) weights/network, (b) weights/layer, (c) activations/network, and (d) activations/layer.}
\label{fig:halves_vis_comparison}
\end{figure*}

\textbf{Halves-MNIST Dataset:}
Figure~\ref{fig:halves_vis_comparison} shows examples from our halves-MNIST dataset. To create each example, two images were randomly selected, resized to have half their original width, and concatenated together. Each image was labeled with the sum of the two digits modulo 10. Examples are shown in Figure \ref{fig:halves_vis}.

\textbf{Visualization:}
To create the images from Figure \ref{fig:halves_vis} which show clusters of neurons in the first layer of an MLP trained on halves-MNIST, we use a correlation-based visualization algorithm from \citet{watanabe2019interpreting}. 
We construct visualizations of neurons using their correlations with the input pixels' pre ReLU activities across the test dataset.
Instead of Pearson (linear) correlation, we use the Spearman correlation (which is the Pearson correlation of ranks) because it how well a relationship monotonically increases even if it is nonlinear.

After obtaining visualizations for each neuron in a subcluster, we do not directly take their average to visualize the entire subcluster. 
To see why, consider two neurons which are highly anticorrelated across the testing set.
These neurons are highly coherent, but averaging together their visualizations would obscure this by cancellation. 
To fix this problem, we align the signs of the visualizations for individual neurons using a variant of a stoahcstic alignment algorithm from \citet{watanabe2019interpreting}. 
To visualize a subcluster, for a number of iterations (we use 20), we iterate over its neurons and calculate for each the sum of cosines between its visualization and each of the other neurons' visualizations in vector form. 
If this sum is negative, we flip the sign of this neuron's visualization. 
After this procedure, we take the mean of the visualizations within a subcluster. 
This process is detailed in Algorithm \ref{alg:sign_slignment}.

\begin{algorithm}[]
\SetAlgoLined
\KwResult{Set of sign-aligned neuron visualizations.}
 \textbf{Input} Neuron visualizations $V_{1:n}$
 \For{iter in num\_iters}{
 \For{$v_i$ in $V$}{
 Calculate sum of cosines, $c = \sum_{j \ne i} \frac{v_i \cdot v_j}{\sqrt{v_i \cdot v_i}\sqrt{v_j \cdot v_j}}$
 
 \If{$c < 0$}{
   $v_i \gets -v_i$
   }
  }
 }
 \caption{Sign Alignment Algorithm (Similar to \citet{watanabe2019interpreting})}
 \label{alg:sign_slignment}
\end{algorithm}

\textbf{Comparison to Random Subclusters:}
To confirm that the visualization in Figure \ref{fig:halves_vis} show that our clustering algorithms are capturing local specialization in the first layer of the MLPs, we compare them to the same visualizations but for random subclusters in Figure \ref{fig:halves_vis_comparison}. 
The visualizations that were used to produce Figure \ref{fig:halves_vis} are each at the top of a column of one of the panels in Figure \ref{fig:halves_vis_comparison} and are above visualizations for random subclusters of the same size and layer. 
Each column in each panel was independently scaled to have values in the interval [0,1].
In each of the four panels, visualizations in the first row reflect the most selectivity to one side or the other meaning that compared to random subclusters, the ones found via these clustering methods tend to be significantly more coherent.

\textbf{MLPs can Compute Modular Sums:} 
A network an do this for $M$ values by using an intermediate layer of $M^2$ neurons, each of which serve as a detector of one of the possible combinations of inputs. 
Consider a ReLU MLP with $2M$ inputs, a single hidden layer with $M^2$ neurons, and then $M$ outputs. 
Suppose that it is given the task of mapping datapoints in which the input nodes numbered $i$ and $M+j$ are activated with value 1 to an output in which the
$(i + j)$\textsuperscript{th} node modulo $M$
is active with value 1. 
It could do so if each hidden neuron with a ReLU activation detected one of the $M^2$ possible input combinations via a bias of -1 and two weights of 1 connecting it to each of the input nodes in the combination is detects. 
A single weight from each hidden neuron to its corresponding output point would allow the network to compute the modular sum. 
In our MLPs, we have $M=10$ classes, and the MLPs have a dense layer with $>10^2$ neurons preceding the output layer. 
Thus, they are capable of computing a modular sum in the halves and stack-diff tasks we give to them. 

\subsection{Feature Visualization} \label{app:feature_visualization}

\begin{figure*}[h!]
\centering
\begin{tabular}{lcc}
    (a) & \includegraphics[width=0.09\columnwidth]{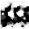} &
    \includegraphics[width=0.81\columnwidth]{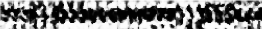} \\
    (b) & \includegraphics[width=0.09\columnwidth]{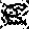} &
    \includegraphics[width=0.81\columnwidth]{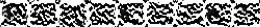} \\
    (c) & \includegraphics[width=0.09\columnwidth]{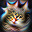} &
    \includegraphics[width=0.81\columnwidth]{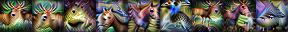} \\
    (d) & \includegraphics[width=0.09\columnwidth]{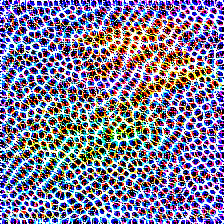} &
    \includegraphics[width=0.81\columnwidth]{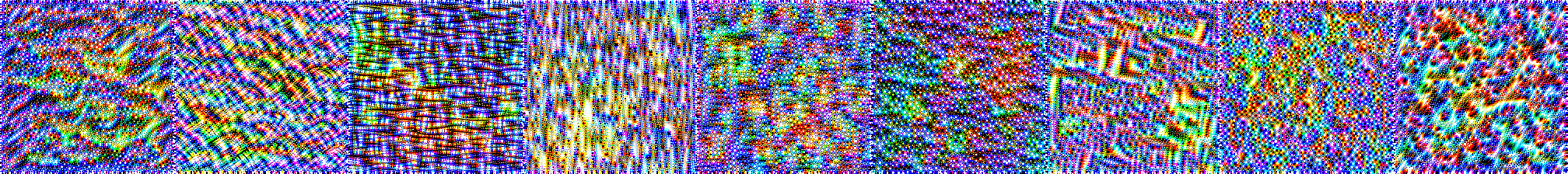} \\
    (e) & \includegraphics[width=0.09\columnwidth]{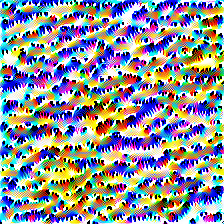} &
    \includegraphics[width=0.81\columnwidth]{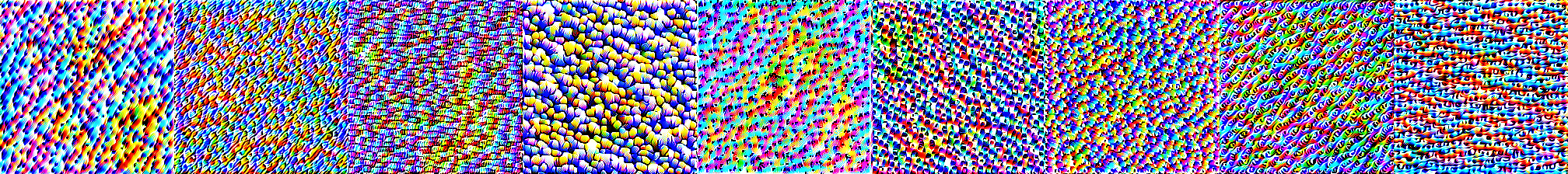} \\
\end{tabular}
\caption{\textbf{Example feature visualizations for true and random sub-clusters:} In the left column are shown true sub-cluster visualizations, and in the right column are visualizations of sub-clusters of random neurons of the same size in the same layer. (a) MLP, MNIST; (b) CNN, MNIST; (c) CNN-VGG, CIFAR-10; (d) VGG-16, ImageNet; (e) VGG-19, ImageNet.
}
\label{fig:lucid_examples}
\end{figure*}

All visualizations were created using the Lucid\footnote{\url{https://github.com/tensorflow/lucid}} package. 
The optimization objective for visualizing sub-clusters was the $L_1$ norm of the pre-ReLU inputs for all neurons inside the subcluster (it was an $L_1$ norm of $L_1$ norms for convolutional feature maps). 
For small MLPs, small CNNs, and mid-sized CNNs, we generated images using random jittering and scaling, and for ImageNet models, we used Lucid’s default transformations which consist of padding, jittering, rotation, and scaling with default hyperparameters. 
For all networks, we used the standard pixel-based parameterization of the image and no regularization on the Adam optimizer for 100 steps. 
For visualizations in small MLPs and CNNs, we used versions of these networks trained on 3-channel versions of their datasets in which the same inputs were stacked thrice because Lucid requires networks to have 3-channel inputs. 
However, we show grayscaled versions of these in figure \ref{fig:lucid_examples}.
Refer to Section \ref{sec:experiments} of the main paper for quantitative analysis of the optimization objective values.

\subsection{Robustness to the Choice of Cluster Number} \label{app:alternate_k}

In the main paper, we only present results form experiments in which the number of clusters, $k$, was set to 16 for sub-ImageNet networks and 32 for ImageNet ones. The fact that we find evidence of local specialization in networks across a range of sizes using $k=16$ suggests that detecting it is robust to $k$. However, here we also present a direct comparison between results for CIFAR CNN-VGGs for $k \in \{8, 12, 16\}$. Tables \ref{tab:lesion_altk} and \ref{tab:lucid_altk} show these results for lesion and feature-visualization experiments respectively. In general, whether or not an experiment resulted in a Fisher statistic or an effect measure suggesting local specialization is consistent across these values of $k$.

\begin{table}[h!]
\footnotesize
\caption{Comparison for different cluster number values in CNN-VGG networks for lesion-based experiments. Results are shown for $k \in \{8, 12, 16\}$. Fisher statistics are means over 5 runs. Each Fisher statistic which is significant at the $\alpha=0.05$ after Benjamini-Hochberg correction is \textbf{bold}. An effect measure $>1$ corresponds to more importance/coherence among true subclusters than random ones. All above 1 are \textbf{bold}.}
\label{tab:lesion_altk}
\begin{center}
\begin{tabular}{l l r r r r }

\toprule
 & &  \multicolumn{2}{c}{\textcolor{blue}{\textbf{Importance: Acc. Drop}}} & \multicolumn{2}{c}{\textcolor{orange}{\textbf{Coherence: Class Range}}}\\
\textbf{Network} & \textbf{Partitioning} & {\textbf{Fisher}} & {\textbf{Effect Meas.}}  & {\textbf{Fisher}} & {\textbf{Effect Meas.}}  \\
\textbf{Cluster number} ($k$) & & \textbf{stat.} & \textbf{High$\to$Imp.} & \textbf{stat.} & \textbf{High$\to$Coh.}  \\

\midrule

& Weight/Network & $\mathbf{1.85}$ & $0.910 \pm 0.102$ & $\mathbf{1.94}$ & $0.270 \pm 0.045$ \\
VGG, CIFAR-10 & Weight/Layer & $1.03$ & $0.854 \pm 0.047$ & $\mathbf{1.17}$ & $0.828 \pm 0.039$ \\
$k=8$ & Act./Network & $\mathbf{1.49}$ & $0.966 \pm 0.038$ & $0.98$ & $0.748 \pm 0.030$ \\
& Act./Layer & $\mathbf{1.77}$ & $1.051 \pm 0.035$ & $0.93$ & $0.752 \pm 0.027$ \\ \hline

& Weight/Network & $\mathbf{1.63}$ & $0.737 \pm 0.078$ & $\mathbf{2.01}$ & $0.388 \pm 0.045$ \\
VGG, CIFAR-10 & Weight/Layer & $1.05$ & $0.867 \pm 0.042$ & $1.09$ & $0.736 \pm 0.035$ \\
$k=12$ & Act./Network & $\mathbf{1.44}$ & $0.901 \pm 0.035$ & $1.04$ & $0.721 \pm 0.028$ \\
& Act./Layer & $\mathbf{1.57}$ & $0.994 \pm 0.032$ & $0.96$ & $0.694 \pm 0.023$ \\ \hline

& Weight/Network & $\mathbf{1.50}$ & $0.682 \pm 0.066$ & $\mathbf{2.12}$ & $0.407 \pm 0.042$ \\
VGG, CIFAR-10 & Weight/Layer & $\mathbf{1.15}$ & $0.808 \pm 0.041$ & $\mathbf{1.27}$ & $0.692 \pm 0.033$ \\
$k=16$ & Act./Network & $\mathbf{1.40}$ & $0.926 \pm 0.032$ & $0.97$ & $0.679 \pm 0.023$ \\
& Act./Layer & $\mathbf{1.56}$ & $0.956 \pm 0.030$ & $1.03$ & $0.695 \pm 0.021$ \\

\bottomrule
\end{tabular}
\end{center}
\end{table}

\begin{table}[h!]
\footnotesize
\caption{Comparison for different cluster values in CNN-VGG networks for feature visualization-based experiments. Results are shown for $k \in \{8,12,16\}$. Fisher statistics are means over 5 runs. Each mean Fisher statistic which is significant at the $\alpha=0.05$ level after Benjamini-Hochberg correction is \textbf{bold}. For vis score tests, an effect measure $>1$ corresponds to more coherence, and for softmax $H$ tests, one of $<1$ corresponds to more coherence. All effect measures on the side of 1 indicating more coherence are \textbf{bold}.}
\label{tab:lucid_altk}
\begin{center}
\begin{tabular}{l l r r r r}
\toprule
 & &  \multicolumn{2}{c}{\textcolor{orange}{\textbf{Coherence: Vis Score}}} & \multicolumn{2}{c}{\textcolor{orange}{\textbf{Coherence: Softmax $\mathbf{H}$}}}\\
\textbf{Network} & \textbf{Partitioning} & {\textbf{Fisher}} & {\textbf{Effect Meas.}}  & {\textbf{Fisher}} & {\textbf{Effect Meas.}}  \\
\textbf{Cluster number} ($k$) & & \textbf{stat.} & \textbf{High$\to$Coh.} & \textbf{stat.} & \textbf{Low$\to$Coh.}  \\
\midrule

& Weight/Network & $\mathbf{1.87}$ & $1.011 \pm 0.016$ & $1.18$ & $1.028 \pm 0.017$ \\
VGG, CIFAR-10 & Weight/Layer & $0.83$ & $0.974 \pm 0.005$ & $0.90$ & $1.059 \pm 0.015$ \\
$k=8$ & Act./Network & $\mathbf{2.61}$ & $\mathbf{1.345 \pm 0.006}$ & $1.02$ & $\mathbf{0.940 \pm 0.010}$ \\
& Act./Layer & $\mathbf{2.72}$ & $\mathbf{1.358 \pm 0.005}$ & $1.06$ & $\mathbf{0.946 \pm 0.009}$ \\ \hline

& Weight/Network & $\mathbf{1.59}$ & $0.959 \pm 0.013$ & $0.89$ & $1.106 \pm 0.013$ \\
VGG, CIFAR-10 & Weight/Layer & $0.96$ & $0.993 \pm 0.005$ & $1.12$ & $\mathbf{0.945 \pm 0.012}$ \\
$k=12$ & Act./Network & $\mathbf{2.60}$ & $\mathbf{1.363 \pm 0.005}$ & $1.12$ & $\mathbf{0.917 \pm 0.009}$ \\
& Act./Layer & $\mathbf{2.73}$ & $\mathbf{1.364 \pm 0.004}$ & $1.07$ & $\mathbf{0.928 \pm 0.008}$ \\ \hline

& Weight/Network & $\mathbf{1.46}$ & $0.871 \pm 0.011$ & $0.99$ & $1.124 \pm 0.012$ \\
VGG, CIFAR-10 & Weight/Layer & $1.00$ & $\mathbf{1.013 \pm 0.006}$ & $0.97$ & $0.992 \pm 0.012$ \\
$k=16$ & Act./Network & $\mathbf{2.34}$ & $\mathbf{1.327 \pm 0.005}$ & $1.08$ & $\mathbf{0.950 \pm 0.009}$ \\
& Act./Layer & $\mathbf{2.67}$ & $\mathbf{1.379 \pm 0.004}$ & $\mathbf{1.12}$ & $\mathbf{0.930 \pm 0.008}$ \\ 

\bottomrule
\end{tabular}
\end{center}
\end{table}

\subsection{Multiple Testing Correction} \label{app:benjamini_hochberg}

In Tables~\ref{tab:lesion} and \ref{tab:lucid}, we report various $p$ values that summarize the degree to which statistics of sub-clusters vary from those of random groups of neurons within a network. 
For each network, the $p$ values one-sidedly test whether the true sub-cluster measures reflect more importance or coherence than those of random subclusters. 
However, when testing multiple networks, one may want to reduce the chance of false positives due to the sheer number of tests performed. 
To incorporate this analysis into our results, we perform a multiple testing correction on Tables~\ref{tab:lesion} and \ref{tab:lucid} using the Benjamini-Hochberg procedure \citep{benjamini1995controlling}. 
This procedure controls the false discovery rate: that is, the expected proportion of rejections of the null hypothesis that are false positives, where the expectation is taken under the data-generating distribution.
It relies on all experiments being independent.
For a false discovery rate $\alpha$ and $m$ ordered $p$ values $\{p_1,\dotsc,p_m\}$, this procedure chooses a critical $p$ value as $p_k$ where $k = \argmax_{j \in 1:m} \mathbb{I}(p_j \le \frac{j \alpha}{m})$ where $\mathbb{I}$ is an indicator.
All $p$ values greater than $p_k$ are deemed not significant at this level. 
In Table~\ref{tab:fisher_stats} we bold Fisher statistics significant at the $\alpha=0.05$ level under this correction.
In our case, the critical value was $p_k = 0.025$.

\subsection{Full Tabular Data} \label{app:tabular_data}

Tables~\ref{tab:lesion} (lesion experiments data) and \ref{tab:lucid} (feature visualization experiment data) show the mean Fisher statistics from Table~\ref{tab:fisher_stats}, their $p$ values, and effect measure results from Table~\ref{tab:effect_sizes}, but do so in a form that places the mean Fisher statistic, $p$ value, and effect measure data from each network in the same row. All Fisher statistics significant according to the Benjamini-Hochberg method are in \textbf{bold}, and all effect measures indicating greater average importance/coherence among true subclusters compared to random subclusters are in \textbf{bold}.

\begin{table}[h!]
\footnotesize
\caption{Results for lesion-based experiments in networks involving importance as measured through overall accuracy drops and coherence as measured by the class-wise range of accuracy drops. Each row corresponds to a network paired with a partitioning method. Results are calculated as explained in Section \ref{sec:statistical_analysis}---in particular, Fisher statistics are means over 5 runs. Each Fisher statistic which is significant at an $\alpha=0.05$ level under the Benjamini Hochberg multiple correction is in \textbf{bold}. In this case, significance means $p \le 0.025$. For both accuracy drop and classwise range experiments, an effect measure $>1$ corresponds to more importance/coherence among true subclusters than random ones. All such effect measures which are further than two standard errors above 1 are in \textbf{bold}.}
\label{tab:lesion}
\begin{center}

\begin{tabular}{l l r r r r }

\toprule
 & &  \multicolumn{2}{c}{\textcolor{blue}{\textbf{Importance: Acc. Drop}}} & \multicolumn{2}{c}{\textcolor{orange}{\textbf{Coherence: Class Range}}}\\
\textbf{Network} & \textbf{Partitioning} & {\textbf{Fisher Stat.}} & {\textbf{Effect Meas.}}  & {\textbf{Fisher Stat.}} & {\textbf{Effect Meas.}}  \\
& & ($p$ Value) & \textbf{High$\to$Imp.} & ($p$ Value) & \textbf{High$\to$Coh.}  \\

\midrule

\multirow{4}{*}{MLP, MNIST} & Weight/Network & $\mathbf{2.13}\; (1 \times 10^{-17})$ & $\mathbf{1.123\pm0.058}$ & $0.91\; (0.82)$ & $0.701\pm0.036$ \\
& Weight/Layer & $\mathbf{2.05}\; (5 \times 10^{-27})$ & ${1.061\pm0.048}$ & $0.91\; (0.91)$ & $0.676\pm0.029$ \\
& Act./Network & $\mathbf{1.46}\; (2 \times 10^{-11})$ & $0.883\pm0.038$ & $1.00\; (0.56)$ & $0.646\pm0.024$ \\
& Act./Layer & $\mathbf{1.69}\; (6 \times 10^{-22})$ & $0.929\pm0.040$ & $1.03\; (0.32)$ & $0.687\pm0.025$ \\ \hline

\multirow{4}{*}{CNN, MNIST} & Weight/Network & $\mathbf{1.29}\; (2 \times 10^{-5})$ & $0.837\pm0.048$ & $0.84\; (0.98)$ & $0.527\pm0.031$ \\
& Weight/Layer & $1.10\; (0.063)$ & $0.814\pm0.046$ & $0.94\; (0.82)$ & $0.635\pm0.033$ \\
& Act./Network & $\mathbf{1.73}\; (4 \times 10^{-10})$ & ${1.078\pm0.061}$ & $0.70\; (1.00)$ & $0.543\pm0.039$ \\
& Act./Layer & $\mathbf{1.46}\; (2 \times 10^{-7})$ & $0.939\pm0.060$ & $0.92\; (0.84)$ & $0.625\pm0.043$ \\ \hline

\multirow{4}{*}{VGG, CIFAR-10} & Weight/Network & $\mathbf{1.50}\; (4 \times 10^{-8})$ & $0.682\pm0.066$ & $\mathbf{2.12}\; (2 \times 10^{-29})$ & $0.407\pm0.042$ \\
& Weight/Layer & $\mathbf{1.15}\; (0.021)$ & $0.808\pm0.041$ & $\mathbf{1.27}\; (2 \times 10^{-4})$ & $0.692\pm0.033$ \\
& Act./Network & $\mathbf{1.40}\; (2 \times 10^{-11})$ & $0.926\pm0.032$ & $0.97\; (0.85)$ & $0.679\pm0.023$ \\
& Act./Layer & $\mathbf{1.56}\; (2 \times 10^{-23})$ & $0.956\pm0.030$ & $1.03\; (0.27)$ & $0.695\pm0.021$ \\ \hline

\multirow{4}{*}{VGG16, ImageNet} & Weight/Network & $\mathbf{2.54}\; (3 \times 10^{-23})$ & $\mathbf{1.205\pm0.050}$ & $0.49\; (1.00)$ & $0.790\pm0.010$ \\
& Weight/Layer & $\mathbf{2.15}\; (2 \times 10^{-57})$ & $\mathbf{1.168 \pm 0.019}$ & $0.56\; (1.00)$ & $0.825\pm0.005$ \\
& Act./Network & $\mathbf{1.89}\; (1 \times 10^{-19})$ & $\mathbf{1.129	\pm 0.026}$ & $0.63\; (1.00)$ & $0.859\pm0.006$ \\
& Act./Layer & $\mathbf{1.66}\; (2 \times 10^{-24})$ & $\mathbf{1.063\pm0.021}$ & $0.70\; (1.00)$ & $0.876\pm0.005$ \\ \hline 

\multirow{4}{*}{ResNet18, ImageNet} & Weight/Network & $\mathbf{1.42}\; (3 \times 10^{-4})$ & $0.926\pm0.045$ & $1.13\; (0.13)$ & $0.957\pm0.011$ \\
& Weight/Layer & $\mathbf{1.29}\; (4 \times 10^{-9})$ & $0.971\pm0.015$ & $0.99\; (0.55)$ & $0.971\pm0.004$ \\
& Act./Network & $\mathbf{1.30}\; (6 \times 10^{-7})$ & $0.979\pm0.017$ & $0.92\; (0.92)$ & $0.977\pm0.004$ \\
& Act./Layer & $\mathbf{1.31}\; (5 \times 10^{-10})$ & $0.971\pm0.015$ & $0.96\; (0.794)$ & $0.967\pm0.004$ \\ 

\bottomrule
\end{tabular}
\end{center}
\end{table}

\begin{table}[h!]
\footnotesize
\caption{Results for feature visualization-based experiments in networks involving coherence as measured by the optimization score of feature visualizations and the entropy of network outputs. Each row corresponds to a network paired with a partitioning method. Results are calculated as explained in Section \ref{sec:statistical_analysis}---in particular, Fisher statistics are means over 5 runs. Each Fisher statistic which is significant at an $\alpha=0.05$ level after Benjamini Hochberg multiple correction is in \textbf{bold}. In this case, significance means $p \le 0.025$. For visualization score experiments, an effect measure $>1$ corresponds to more coherence among true subclusters than random ones, while one of $<1$ does for softmax entropy experiments. All effect measures which are more than two standard errors away from 1 on the side reflecting greater coherence among true subclusters are in \textbf{bold}}
\label{tab:lucid}
\begin{center}
\begin{tabular}{l l r r r r}
\toprule
 & &  \multicolumn{2}{c}{\textcolor{orange}{\textbf{Coherence: Vis Score}}} & \multicolumn{2}{c}{\textcolor{orange}{\textbf{Coherence: Softmax $\mathbf{H}$}}}\\
\textbf{Network} & \textbf{Partitioning} & {\textbf{Fisher Stat.}} & {\textbf{Effect Meas.}}  & {\textbf{Fisher Stat.}} & {\textbf{Effect Meas.}}  \\
& & ($p$ Value) & \textbf{High$\to$Coh.} & ($p$ Value) & \textbf{Low$\to$Coh.}  \\
\midrule

\multirow{4}{*}{MLP, MNIST} & Weight/Network & $\mathbf{1.32}\; (0.001)$ & $1.003\pm0.004$ & $0.92\; (0.71)$ & $1.105\pm0.021$ \\
& Weight/Layer & $\mathbf{1.21}\; (0.002)$ & $\mathbf{1.024\pm0.004}$ & $\mathbf{1.23}\; (0.023)$ & $0.931\pm0.016$ \\
& Act./Network & $\mathbf{1.34}\; (4 \times 10^{-5})$ & $\mathbf{1.020\pm0.003}$ & $\mathbf{1.15}\; (0.025)$ & ${0.997\pm0.013}$ \\
& Act./Layer & $\mathbf{1.36}\; (9 \times 10^{-7})$ & $\mathbf{1.026\pm0.003}$ & $\mathbf{1.12}\; (0.013)$ & ${1.011\pm0.012}$ \\
\hline

\multirow{4}{*}{CNN, MNIST} & Weight/Network & $1.10\; (0.082)$ & $0.998\pm0.004$ & $0.93\; (0.86)$ & $1.026\pm0.008$ \\
& Weight/Layer & $1.02\; (0.34)$ & ${1.004\pm0.003}$ & $0.99\; (0.52)$ & $1.007\pm0.007$ \\
& Act./Network & $1.09\; (0.27)$ & $0.933\pm0.006$ & $0.90\; (0.83)$ & $1.025\pm0.011$ \\
& Act./Layer & $1.05\; (0.22)$ & $0.925\pm0.005$ & $0.98\; (0.67)$ & $\mathbf{0.970\pm0.010}$ \\
\hline

\multirow{4}{*}{VGG, CIFAR-10} & Weight/Network & $\mathbf{1.46}\; (7 \times 10^{-8})$ & $0.871\pm0.011$ & $0.99\; (0.61)$ & $1.124\pm0.012$ \\
& Weight/Layer & $1.00\; (0.44)$ & $\mathbf{1.013\pm0.006}$ & $0.97\; (0.62)$ & $0.992\pm0.012$ \\
& Act./Network & $\mathbf{2.34}\; (6 \times 10^{-64})$ & $\mathbf{1.327\pm0.005}$ & $1.08\; (0.045)$ & $\mathbf{0.950\pm0.009}$ \\
& Act./Layer & $\mathbf{2.67}\; (8 \times 10^{-121})$ & $\mathbf{1.379\pm0.004}$ & $\mathbf{1.12}\; (0.013)$ & $\mathbf{0.930\pm0.008}$ \\ \hline

\multirow{4}{*}{VGG-16, ImageNet} & Weight/Network & $\mathbf{1.72}\; (6 \times 10^{-8})$ & $\mathbf{1.043\pm0.005}$ & $1.19\; (0.052)$ & ${0.998\pm0.001}$ \\
& Weight/Layer & $\mathbf{1.90}\; (3 \times 10^{-39})$ & $\mathbf{1.076\pm0.003}$ & $1.06\; (0.16)$ & $\mathbf{0.991\pm0.001}$ \\
& Act./Network & $\mathbf{1.83}\; (3 \times 10^{-15})$ & $\mathbf{1.066\pm0.003}$ & $1.07\; (0.21)$ & $1.000\pm0.001$ \\
& Act./Layer & $\mathbf{1.85}\; (1 \times 10^{-34})$ & $\mathbf{1.056\pm0.003}$ & $0.98\; (0.63)$ & $1.001\pm0.001$ \\ \hline

\multirow{4}{*}{VGG-19, ImageNet} & Weight/Network & $\mathbf{1.91}\; (2 \times 10^{-10})$ & $\mathbf{1.061\pm0.004}$ & $1.03\; (0.40)$ & $1.003\pm0.001$ \\
& Weight/Layer & $\mathbf{2.23}\; (4 \times 10^{-81})$ & $\mathbf{1.099\pm0.003}$ & $1.00\; (0.46)$ & $1.001\pm0.001$ \\
& Act./Network & $\mathbf{1.87}\; (6 \times 10^{-22})$ & $\mathbf{1.046\pm0.003}$ & $1.10\; (0.089)$ & $\mathbf{0.996\pm0.001}$ \\
& Act./Layer & $\mathbf{2.01}\; (3 \times 10^{-27})$ & $\mathbf{1.081\pm0.002}$ & $0.98\; (0.65)$ & $1.004\pm0.001$ \\

\bottomrule
\end{tabular}
\end{center}
\end{table}

\end{document}